\documentclass[10pt,twocolumn,letterpaper]{article}

\usepackage{cvpr}              
\usepackage[dvipsnames]{xcolor}

\usepackage{graphicx}
\usepackage{cite}
\usepackage{color}
\usepackage{wrapfig}
\usepackage{epsfig}
\usepackage{graphicx}
\usepackage{amsmath}
\usepackage{amssymb}
\usepackage{multirow}
\usepackage{algorithmic}
\usepackage{arydshln}
\usepackage{threeparttable}
\usepackage{colortbl}
\usepackage{enumitem}
\usepackage{siunitx}
\usepackage{bm}
\usepackage{array}
\usepackage{color}
\usepackage{colortbl}
\usepackage{caption}
\usepackage{dblfloatfix}
\usepackage{amssymb}
\usepackage{pifont}
\usepackage{amssymb}
\usepackage{float}

\usepackage{listings}
\usepackage[export]{adjustbox}
\usepackage{makecell}
\usepackage{tikz}
\newcommand{\aligntridown}{\raise.4ex\hbox{$\bigtriangledown$}}

\usepackage{subcaption}
\usepackage{dsfont}
\definecolor{mygray}{gray}{.9}
\definecolor{ggray}{RGB}{127,127,127}
\definecolor{reda}{RGB}{192,0,0}
\definecolor{redb}{RGB}{217,148,143}
\definecolor{myyellow}{RGB}{190,144,0}
\definecolor{mygreen}{RGB}{80,100,40}
\definecolor{myblue}{RGB}{30,90,100}

\newcommand{\revise}[1]{{\color{black} #1}}

\newcolumntype{x}[1]{>{\centering\arraybackslash}p{#1pt}}

\makeatletter\renewcommand\paragraph{\@startsection{paragraph}{4}{\z@}
    {.5em \@plus1ex \@minus.2ex}{-.5em}{\normalfont\normalsize\bfseries}}\makeatother
\usepackage{siunitx}
\usepackage[nopar]{lipsum}
\usepackage[export]{adjustbox}

\def\ms{$_{\!}$ }

\makeatletter
\newcommand{\thickhline}{%
    \noalign {\ifnum 0=`}\fi \hrule height 1pt
    \futurelet \reserved@a \@xhline
}

\makeatother

\usepackage{caption}
\usepackage[utf8]{inputenc}

\definecolor{bblue}{RGB}{0,30,95}
\definecolor{rred}{RGB}{190,0,0}
\definecolor{ggray}{RGB}{127,127,127}
\usepackage[linesnumbered,ruled]{algorithm2e}

\definecolor{cvprblue}{rgb}{0.21,0.49,0.74}
\usepackage[pagebackref,breaklinks,colorlinks,urlcolor=gray,citecolor=cvprblue]{hyperref}

\def\confName{CVPR}
\def\confYear{2024}

\title{Clustering for Protein Representation Learning}

\author{Ruijie Quan,~ Wenguan Wang\thanks{Corresponding author: \textit{Wenguan Wang}.},~ Fan Ma,~ Hehe Fan,~ Yi Yang~\\
\small{ReLER,~ CCAI,~ Zhejiang University}\\\small\url{https://github.com/QUANRJ/ClusteringPRL}}

\begin{document}
\maketitle
\begin{abstract}
Protein representation learning is a challenging task that aims to capture the structure and function of proteins from their amino acid sequences. Previous methods largely ignored the fact that not all amino acids are equally important for protein folding and activity. In this article, we propose a neural clustering framework that can automatically discover the critical components of a protein by considering both its primary and tertiary structure information. Our framework treats a protein as a graph, where each node represents an amino acid and each edge represents a spatial or sequential connection between amino acids. We then apply an iterative clustering strategy to group the nodes into clusters based on their 1D and 3D positions and assign scores to each cluster. We select the highest-scoring clusters and use their medoid nodes for the next iteration of clustering, until we obtain a hierarchical and informative representation of the protein. We evaluate on four protein-related tasks: protein fold classification, enzyme reaction classification, gene ontology term prediction, and enzyme commission number prediction. Experimental results demonstrate that our method achieves state-of-the-art performance. 
\end{abstract}    
\section{Introduction}
\label{sec:intro}

Proteins are one of the most fundamental elements in living organisms and make significant contributions to nearly all fundamental biological processes in the cell. Composed of one or several chains of amino acids~\cite{sanger1951amino,pauling1951structure}, proteins fold into specific conformations to enable various biological functionalities~\cite{govindarajan1999estimating}. A multi-level structure of proteins begins with the \textit{primary structure}, which is defined by the sequence of amino acids forming the protein backbone~\cite{sanger1952arrangement}. The \textit{secondary structure} is determined by hydrogen bonds between distant amino acids in the chain, resulting in substructures such as $\alpha$-helices and $\beta$-sheets~\cite{sun2004overview}. \textit{Tertiary structure} arises from folding of secondary structures, determined by interactions between side chains of amino acids~\cite{branden2012introduction}. Lastly, the \textit{quarternary structure} describes the arrangement of polypeptide chains in a multi-subunit arrangement~\cite{chou2003predicting}. Understanding the structure and function of proteins~\cite{gligorijevic2021structure,hermosilla2020intrinsic,zhang2022protein,wanglearning2023,fancontinuous,Fan_2023_CVPR,wang2024protchatgpt} is crucial in elucidating their role in biological processes and developing new therapies and treatments for a variety of diseases.

While a protein's conformation and function are primarily determined by its amino acid sequence, it is important to recognize that not all amino acids contribute equally to these aspects.
In fact, certain amino acids, known as the \textit{critical components}, play the primary role in determining a protein's shape and function~\cite{ingram2004sickle,noguchi1981intracellular,orth1979adrenocorticotropic,wheatland2004molecular,frazier1972nerve,blundell1975evolution,doolittle1981similar}. Sometimes, even a single amino acid substitution can significantly impact a protein's overall structure and function~\cite{ingram2004sickle,noguchi1981intracellular}. For example, sickle cell anemia results from a single amino acid change in hemoglobin, causing it to form abnormal fibers that distort red blood cell shape. Besides, the critical components of a protein's primary structure are essential for its biological activity. For instance, any of the first 24 amino acids of the adrenocorticotropic hormone (ACTH) molecule is necessary for its biological activity, whereas removing all amino acids between 25-39 has no impact~\cite{orth1979adrenocorticotropic,wheatland2004molecular}. Also, proteins with the same critical components perform the same function, \eg, the insulin A and B chains in various mammals contain 24 invariant amino acid residues necessary for insulin function, while differences in the remaining amino acid residues do not impact insulin function~\cite{frazier1972nerve,blundell1975evolution}. 
In addition, proteins from the same family often have long stretches of similar amino acid sequences within their primary structure~\cite{doolittle1981similar}, suggesting that only a small portion of amino acids that differentiate these proteins are the critical components.

Motivated by the fact that certain amino acids play a more critical role in determining a protein's structure and function than the others~\cite{ingram2004sickle,noguchi1981intracellular,orth1979adrenocorticotropic,wheatland2004molecular,frazier1972nerve,blundell1975evolution,doolittle1981similar}, we devise a neural clustering framework for protein representation learning. 
Concretely, it progressively groups amino acids so as to find the most representative ones for protein classification.
During each iteration, our algorithm proceeds three steps: spherical cluster initialization (SCI), cluster representation extraction (CRE), and cluster nomination (CN).
The iterative procedure starts by treating a protein as a graph where each node represents an amino acid, 
and each edge represents a spatial or sequential connection between amino acids.
In SCI step ($\bigtriangleup$), all the nodes are grouped into clusters based on their sequential and spatial positions. 
Subsequently, in CRE step ($\square$), the information of neighboring nodes within the same cluster are aggregated into a single representative node, namely medoid node. 
This step effectively creates an informative and compact representation for each cluster.
Lastly,\ms in\ms CN\ms step\ms ($\aligntridown$),\ms a\ms graph\ms convolutional\ms network\ms (GCN)~\cite{scarselli2008graph}\ms is\ms applied\ms to\ms score\ms all\ms the\ms clusters,\ms and\ms a\ms few\ms top-scoring\ms ones\ms are\ms selected\ms and\ms their\ms medoid nodes are used as the input for the next iteration.
By iterating these steps $\circlearrowright$($\bigtriangleup\square\aligntridown$), we explore the protein's structure and discover the representative amino acids for protein representation, leading to a powerful, neural clustering based protein representation learning framework.

By$_{\!}$ embracing$_{\!}$ the$_{\!}$ powerful$_{\!}$ idea$_{\!}$ of$_{\!}$ clustering,$_{\!}$ our$_{\!}$ approach$_{\!}$ favorably$_{\!}$ outperforms$_{\!}$ advanced$_{\!}$ competitors. We observe notable improvements of 
\textbf{5.6}\% and \textbf{2.9}\% in \textit{F}\textsubscript{max} for enzyme commission number prediction~\cite{gligorijevic2021structure} (\S\!~\ref{subsec:EC}) and gene ontology term prediction~\cite{gligorijevic2021structure} (\S\!~\ref{subsec:GOTP}). Our method also yields remarkable enhancements of \textbf{3.3}\% and \textbf{1.1}\% in classification accuracy for protein fold classification~\cite{hou2018deepsf} (\S\!~\ref{subsec:PFC}) and enzyme reaction classification~\cite{hermosilla2020intrinsic} (\S\!~\ref{subsec:ERC}). We also provide comprehensive diagnostic analyses (\S\!~\ref{subsec:analysis}) and visual results (\S\!~\ref{sec:visualization}), verifying the efficacy of our essential algorithm designs, showing strong empirical evidence for our core motivation, and confirming the capability of our algorithm in identifying functional motifs of proteins. 
\section{Related Work}
\label{sec:related}

\textbf{Protein Representation Learning}. It has been a topic of interest in the field of bioinformatics and computational biology in recent years. 
Existing methods for this topic can be broadly categorized into two types: \textit{sequence-based} and \textit{structure-based}. 
Early works on sequence-based protein representation learning typically apply word embedding algorithms~\cite{asgari2015continuous,yang2018learned} and 1D convolutional neural networks~\cite{kulmanov2018deepgo,kulmanov2020deepgoplus,hou2018deepsf,tsubaki2019compound}. Though straightforward, these methods neglect the spatial information in protein structures.
To address this limitation, structure-based methods explore the use of 3D convolutional neural networks~\cite{amidi2018enzynet,derevyanko2018deep,townshend2019end} and GCNs~\cite{kipf2016semi,hamilton2017inductive,gligorijevic2021structure,baldassarre2021graphqa,jing2021equivariant,wanglearning2023,zhang2022protein} for this task. 
Recently, some approaches~\cite{jing2021equivariant,wanglearning2023} focus on atom-level representations, treating each atom as a node. The state-of-the-art performance has been achieved by learning at the amino acid-level~\cite{zhang2022protein,fancontinuous}, indicating that protein representation is more closely related to amino acids rather than individual atoms.

Despite significant progress made by these existing methods, they treat all amino acids equally.
In sharp contrast, we propose to learn the protein representation by a neural clustering framework.
This allows us to capture the inherent variability and significance of different amino acids, leading to a more comprehensive and accurate representation of proteins.
We are of the opinion that our method has several potential applications and extensions in the field of protein science. For instance, our neural clustering approach can benefit protein design by utilizing the learned crucial components to direct the design of novel$_{\!}$ protein$_{\!}$ sequences~$_{\!}$~\cite{strokach2020fast,cao2021fold2seq}$_{\!}$ that$_{\!}$ possess$_{\!}$ specific$_{\!}$ properties$_{\!}$ or$_{\!}$ functions.$_{\!}$ This,$_{\!}$ in$_{\!}$ turn,$_{\!}$ can$_{\!}$ facilitate$_{\!}$ the$_{\!}$ development$_{\!}$ of$_{\!}$ new$_{\!}$ therapies$_{\!}$ and$_{\!}$ treatments$_{\!}$ for$_{\!}$ a$_{\!}$ diverse$_{\!}$ range$_{\!}$ of$_{\!}$ illnesses.

\noindent
\textbf{Clustering.} 
Clustering is a fundamental data analysis task that aims to group similar samples together based on similarity, density, intervals or particular statistical distribution measures of the data space~\cite{hartigan1979algorithm, ng2001spectral,janani2019text}. 
It helps to identify representative patterns in data which is meaningful for exploratory data analysis.
Traditional clustering methods~\cite{lloyd1982least,reynolds2009gaussian}
heavily rely on the original data representations.
As a result, they often prove to be ineffective when confronted with data residing in high-dimensional spaces, such as images and text documents.
Recently, deep learning-based clustering methods have attracted increased attention, and been successfully applied to various real-world applications,
such as image segmentation~\cite{xu2022groupvit,liang2023clustseg,zhou2022rethinking,liang2022gmmseg,wang2018semi,ding2024s2vnet}, 
unsupervised representation learning~\cite{wang2022visual,quan2021progressive,zhan2020online,chen2021jigsaw,luo2022clear,yin2022proposalcontrast,feng2023clustering,guikun2024,lu2024zero,feng2024I3D}, 
financial analysis~\cite{close2020combining,govindasamy2018cluster,jaiswal2020green},
and text analysis~\cite{aggarwal2012survey,abualigah2022efficient,subakti2022performance}.

Drawing inspiration from the biological fact that the significance of various amino acids varies, we propose a neural clustering framework for end-to-end protein representation learning. Our objective is to leverage the inherent benefits of clustering to identify the representative amino acids.  
In experiments, we demonstrate the feasibility of our clustering-based method through numerical evaluations and provide visual evidence to reinforce the underlying motivation behind our approach.

\section{Methodology}

\begin{figure*}[t]    
    \centering
    \vspace{-15pt}
    \includegraphics[width=0.8\textwidth]{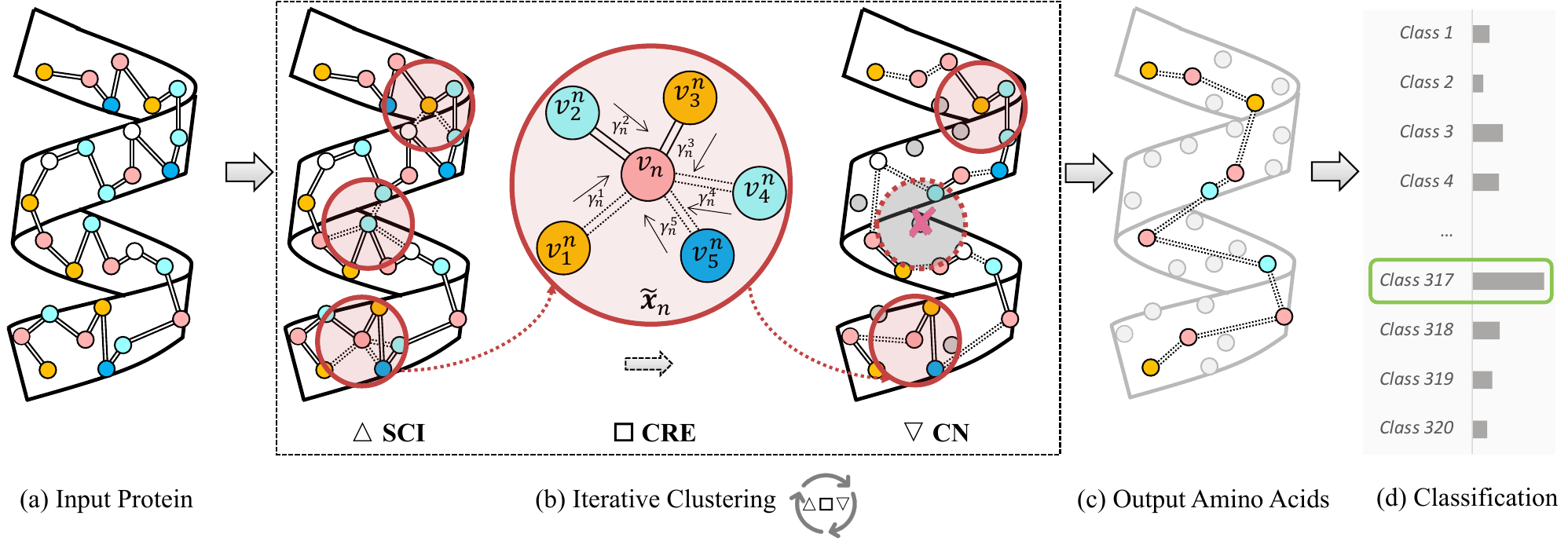}
    \vspace{-14pt}
    \caption{Overview of our iterative neural clustering pipeline for protein representation learning: (a) input protein with amino acids, (b) iterative clustering algorithm which repeatedly stacks three steps $\circlearrowright$($\bigtriangleup\square\aligntridown$), (c) output can be seen as the critical amino acids of the protein, (d) output amino acids used for classification.
    The details of our iterative neural clustering method can be seen in~\S\!~\ref{subsec:hier_clus}.}
        \label{fig:clustering_alg}
    \vspace{-15pt}
\end{figure*}

We have stated that the structure and function of a protein are represented by certain critical amino acids in \S\!~\ref{sec:intro}. This motivates us to regard the protein representation learning task as an amino acid clustering process, so as to automatically explore the critical components of the protein. Firstly, some basic concepts used throughout the paper and the task setup are introduced in \S\!~\ref{sub_sec:nota}. 
Then, we$_{\!}$ elaborate$_{\!}$ on$_{\!}$ the$_{\!}$ neural$_{\!}$ clustering$_{\!}$ framework$_{\!}$ in$_{\!}$ \S\!~\ref{subsec:hier_clus}.$_{\!}$ 
Finally,$_{\!}$ \S\!~\ref{subsec:impl}$_{\!}$ presents$_{\!}$ our$_{\!}$ implementation$_{\!}$ details. 

\subsection{Notation and Task Setup}\label{sub_sec:nota}
A protein is denoted as a triplet $\mathcal{P}=(\mathcal{V},\mathcal{E},\mathcal{Y})$, where $\mathcal{V}=\{v_1,\cdots,v_N\}$ is the set of nodes representing $N$ amino acids, $\mathcal{E}$ the set of edges representing spatial or sequential connections between amino acids, and
$\mathcal{Y}$ the set of labels. The target goal of protein classification is to learn a mapping $\mathcal{V}\rightarrow{\mathcal{Y}}$. Specifically, in single-label classification, \eg, protein fold classification and enzyme reaction classification, the focus is on learning from a collection of examples that are linked to a \textit{single} label from $\mathcal{Y}$. While in multi-class classification, \eg, enzyme commission number prediction and gene ontology term prediction, each examples are associated with \textit{multiple} labels from $\mathcal{Y}$. In what follows, we use $\{\bm{x}_1,\cdots,\bm{x}_N\}$ to denote the features of $\mathcal{V}$, where $\bm{x}_n\in{\mathbb{R}^{256}}$ is the feature vector of amino acid $v_n$. The feature vector can be derived from various sources, such as the one-hot encoding of amino acid types, the orientations of amino acids, and the sequential and spatial positions of amino acids. 
We use $A\in{\{0,1\}^{N\times{N}}}$ to denote the adjacency matrix of $\mathcal{V}$, where $A_{n,m}=1$ if there exists an edge between amino acid nodes $v_n$ and  $v_m$, \ie, $e_{n,m}\in\mathcal{E}$.

\subsection{Iterative Clustering} \label{subsec:hier_clus} 
In our neural clustering framework, we perform iterative clustering on amino acids of the input protein.
Each iteration encompasses three steps, spherical Cluster Initialization (SCI), Cluster Representation Extraction (CRE), and  Cluster Nomination (CN), as illustrated in Figure~\ref{fig:clustering_alg}.

\noindent
\textbf{Spherical Cluster Initialization (SCI).}
 For each amino acid, we initialize a cluster by considering other locally neighboring amino acids within a fixed receptive field, \revise{which draws inspiration from previous work in 3D geometry~\cite{qi2017pointnet}}. 
 This approach enables the examination and comprehension of the spatial and sequential associations among amino acids, which hold paramount significance in determining the structure and functioning of proteins~\cite{fancontinuous}.
Specifically, for each amino acid $v_n$, we define a cluster as a set of amino acids within a fixed radius $r$, where  $v_n$ is regarded as the \textit{medoid} node of the cluster. 
The fixed radius $r$ is a hyperparameter that determines the extent to which the local amino acid nodes are considered for cluster initialization, and its impact on the final performance of protein analysis is studied in \S\!~\ref{subsec:analysis}.
For $v_n$, we denote the set of its neighbor amino acid nodes as $\mathcal{H}^n=\{v^n_1, \cdots, v^n_{K}\}$. Note that $v_n\in{\mathcal{H}^n}$.
 In the first iteration ($t\!=\!1$), we conduct SCI process based on all the input $N$ amino acids to form the clusters. 
In subsequent iterations ($t\!>\!1$), we use the nominated $N_{t-1\!}$ amino acids from the previous $t\!-\!1$-th iteration to initialize the clusters. 
 This allows to focus on exploring the critical components of the protein graph and avoid redundantly exploring the same areas. The adjacency matrix $A$ is regenerated in each SCI process with considering the connectivity among amino acids.

\begin{figure*}[t]
    \centering
    \vspace{-15pt}
    \includegraphics[width=0.85\textwidth]{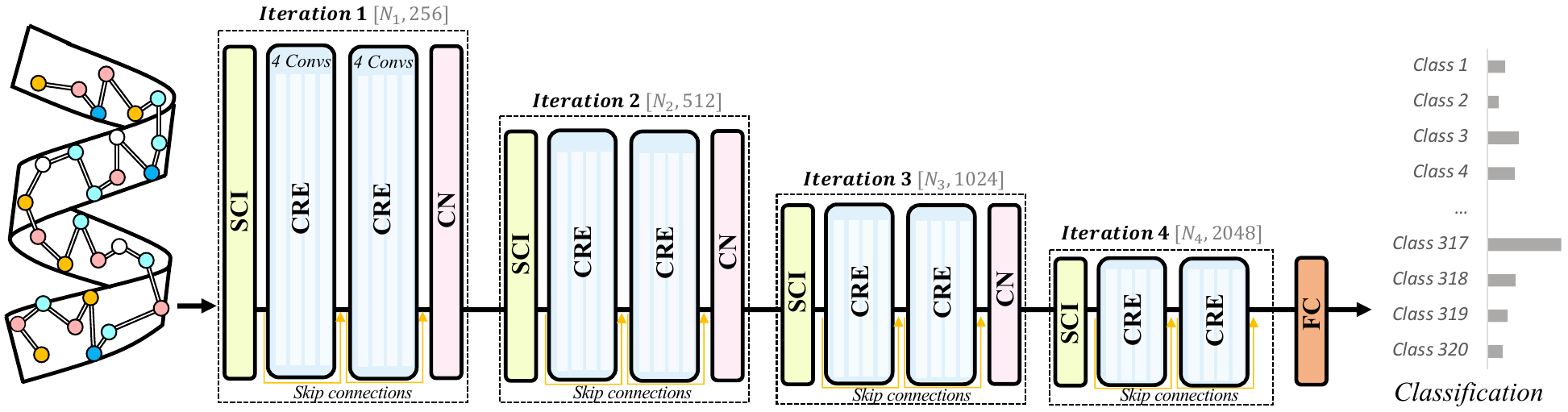}

    \caption{Our neural clustering framework architecture with four iterations. Given a protein, a set of 1D and 3D amino acids, our method adopts an iterative clustering algorithm to explore the most representative amino acids. At each iteration, $B$ cluster representation extraction blocks are utilized to extract cluster features. The clustering nomination operation selects the fraction $\omega$ of amino acids for the next iteration, that ${N_t}\!=\!\lfloor\omega\cdot{N_{t-1}}\rfloor$. Details of the framework can be seen in~\S\!~\ref{subsec:impl}.}
    \vspace{-16pt}
    \label{fig:pipeline}
\end{figure*}

\noindent
\textbf{Cluster Representation Extraction (CRE).}
The second step aims to learn the overall representation of each cluster $\mathcal{H}^n$ by considering all the amino acids within it.
Specifically, we construct the feature representation $\bm{x}_k^n$ of the amino acid node $v_k^n$ in the cluster $\mathcal{H}^n$ by:
\vspace{-5pt}
\begin{equation}\small
 \bm{x}_k^n=f(\bm{g}_k^n, \bm{o}_k^n, {d}_k^n, {s}_k, \bm{e}_k),
 \vspace{-5pt}
 \end{equation}
 where $\bm{g}_k^n\!=\!(\bm{z}_k\!-\!\bm{z}_n)\!\in\!{\mathbb{R}^{3\!}}$ denotes relative geometric coordinates, and $\bm{z}$ indicates the spatial coordinate; 
 $\bm{o}_k^n\!\in\!{\mathbb{R}^{3\!}}$ is the 3D orientation vector; ${d}_k^n\!=\!(||\bm{z}_k\!-\!\bm{z}_n||_2)\!\in\!{\mathbb{R}^{3\!}}$ indicates the spatial distance between $v_n$ and $v^n_k$; ${s}_{k\!}$ is the sequential order on the amino acid chain that is relative to the beginning of the protein sequence; $\bm{e}_k\!\in\!{\mathbb{R}^{128}}$ denotes amino acid embedding (\ie, one-hot encoding of the amino acid type) of  $v_k^n$ in the cluster; and $f$ denotes an encoder which is implemented by a multilayer perceptron. 
In this way, our neural clustering framework considers both primary (1D sequence-based distance) and tertiary (3D coordinates and orientations) structure information of proteins.
Then, we utilize the cross-attention mechanism~\cite{vaswani2017attention} to calculate the attention score $\gamma_k^n\!\in\!(0,1)$ between the medoid amino acid node feature $\bm{x}_n$ and all the constituent ones $\bm{x}_k^n$ in the cluster $\mathcal{H}^n$:
\vspace{-5pt}
\begin{equation}\small
	\gamma_k^n=\frac{\text{exp}(\bm{w}[\bm{x}_n,\bm{x}_k^n])}{\sum\nolimits_{k=1}^{K}\text{exp}(\bm{w}[\bm{x}_n,\bm{x}_k^n])},
\vspace{-5pt}
\end{equation}
where $\bm{w}$ is a learnable vector, and $[\cdot]$ refers to the concatenation operation. 
The attention score $\gamma_k^n$ denotes the level of focus of the medoid on other constituent amino acids.
 Then, the overall cluster representation $\tilde{\bm{x}}_{n}$ for the cluster $\mathcal{H}^n$ of node $v_n$ is given as:
\vspace{-5pt}
\begin{equation}\small
	\tilde{\bm{x}}_{n}=\sum\nolimits_{k=1}^{K}\gamma_k^n\bm{x}_k^n.
\vspace{-5pt}
\end{equation}
\textbf{Cluster Nomination (CN).}
To identify the most representative amino acids (\ie, critical components) in the protein, we propose a cluster nomination process that learns to automatically select these critical components based on the cluster representations $\{\tilde{\bm{x}}_{n}\}_{n=1}^N$. 
Specifically, such cluster nomination process is achieved by a GCN, which takes each cluster representation $\tilde{\bm{x}}_{n}$ as input and calculates a \textit{nomination} score $\varphi_{n}\in(0,1)$:
\vspace{-5pt}
\begin{equation}\small
    \varphi_{n} = \sigma( \bm{W}_{1}\tilde{\bm{x}}_{n} + \sum\nolimits_{m=1}^{N_t} A_{n,m}  (\bm{W}_{2}\tilde{\bm{x}}_{n}  - \bm{W}_{3}\tilde{\bm{x}}_{m} )),
\vspace{-5pt}
\end{equation} 
where $\bm{W}_{1,2,3}$ are learnable parameters and $\sigma$ is ReLU function. 
By utilizing self-loops and the capability to learn functions of local extremas, we are able to score clusters based on both their global and local significance.
The cluster feature is then weighted by the calculated scores: 
\vspace{-5pt}
\begin{equation}\small
	\hat{X}^{c}=\Phi\odot{X^{c}},
\vspace{-5pt}
\end{equation}
where $\Phi\!=\![\varphi_{1},\cdots,\varphi_{N_t}]^{\top}$, $X^{c}\!=\![\tilde{\bm{x}}_{1},\cdots,\tilde{\bm{x}}_{N_t}]$, and $\odot$ is broadcasted hadamard product.  
Based on the weighted cluster features  $\hat{X}^{c}$ and the calculated nomination scores $\Phi$, we select top-${N_t}$ clusters at the $t$-th iteration.
Then, the top-${N_t}$ amino acids are utilized to form a new graph, viewed as the input for the next $t\!+\!1$-th iteration. 
Here ${N_t}$ is the number of selected clusters at $t$-th iteration and is determined by the cluster nomination fraction $\omega$, denoted as
${N_t}\!=\!\lfloor\omega\!\cdot\!{N_{t-1}}\rfloor$, that $t\!\in\!\{1,...,T\}$ and ${N_0}\!=\!N$. We will illustrate the impact of different values of $\omega$ in \S\!~\ref{subsec:analysis}.
\revise{Once the clusters are nominated in the CN phase, the
medoids are then linked following the original sequential orders
of the protein chain.}

At each clustering iteration, we estimate cluster membership and centers by considering the sequential and spatial information of amino acids. Thus it explicitly probes the structures of the portion and captures the complex relationships among amino acids. Then, the representative amino acids (\ie, cluster medoids) are identified and only those most informative ones (which can be viewed as critical components) are selected for next-iteration clustering, and eventually used for functionality prediction. 
After $T$ iterations, a fully connected layer is used to project the feature representations of the nominated $N_T$ amino acids to a $|\mathcal{Y}|$-dimension score vector for classification.
Note that the attention-based selection is implicitly learnt from the supervision signals of the protein classification task.
In a nutshell, our method utilizes an iterative clustering algorithm that repeatedly stacks three steps: SCI, CRE, and CN.

\begin{table*}[!t]
    \centering
    \vspace{-15pt}
    \caption{\textit{F}\textsubscript{max} on EC and GO prediction and Accuracy (\%) on fold and reaction classification. [\textdagger] denotes results taken from~\cite{wang2022lm} and [*] denotes results taken  from~\cite{hermosilla2020intrinsic} and~\cite{anonymous2022contrastive}~(\S\!~\ref{subsec:EC}-\S\!~\ref{subsec:ERC}).}
    \label{tab:all_result}
    \resizebox{\textwidth}{!}{
		\setlength\tabcolsep{6pt}
		\renewcommand\arraystretch{1}
		\footnotesize
        \begin{tabular}{rl||rl||c|ccc|clll|l}
            \thickhline 
            \rowcolor{mygray}\multicolumn{2}{c||}{\multirow{2}{*}{ }} &
            \multicolumn{2}{c||}{\multirow{2}{*}{ }} & 
            \multirow{2}{*}{ }&
            \multicolumn{3}{c|}{ {GO}}&
            \multicolumn{4}{c|}{ {Fold Classification}}&
            \multirow{2}{*}{ }\\
            \cline{6-8}
            \cline{9-12}
            \rowcolor{mygray}\multicolumn{2}{c||}{\multirow{-2}{*}{Method}}& \multicolumn{2}{c||}{\multirow{-2}{*}{Publication}} & \multirow{-2}{*}{EC} & {BP} &  {MF}&  {CC} &  {Fold} &  {Super.} &  {Fam.} &  {Avg.} & \multirow{-2}{*}{Reaction}\\
            \hline \hline

             ResNet&\cite{tape2019} & \textit{NeurIPS}&\textit{2019}
             & 0.605& 0.280& 0.405& 0.304&10.1 & 7.21 & 23.5 & 13.6 & 24.1\\
             LSTM
            &\cite{tape2019} & \textit{NeurIPS}&\textit{2019}
             & 0.425& 0.225& 0.321& 0.283&6.41 & 4.33 & 18.1 & 9.61 & 11.0\\
             Transformer
            &\cite{tape2019} & \textit{NeurIPS}&\textit{2019}
            & 0.238& 0.264& 0.211& 0.405&9.22 & 8.81 & 40.4 & 19.4 & 26.6\\
             GCN
            &\cite{kipf2017semi} & \textit{ICLR}&\textit{2017}
             & 0.320& 0.252& 0.195& 0.329&16.8* & 21.3* & 82.8* & 40.3* & 67.3*\\
             GAT
            &\cite{velickovic2018graph} & \textit{ICLR}&\textit{2018}
             & 0.368& 0.284\textsuperscript{\textdagger}& 0.317\textsuperscript{\textdagger}& 0.385\textsuperscript{\textdagger}&12.4 & 16.5 & 72.7& 33.8 & 55.6\\
             GVP
            &\cite{jing2021equivariant} & \textit{ICLR}&\textit{2021}
             & 0.489& 0.326\textsuperscript{\textdagger}& 0.426\textsuperscript{\textdagger}& 0.420\textsuperscript{\textdagger}&16.0 & 22.5 & 83.8 & 40.7 & 65.5\\
             3DCNN
            &\cite{derevyanko2018deep} & \textit{Bioinform}&\textit{2018}
             & 0.077& 0.240& 0.147& 0.305&31.6* & 45.4* & 92.5* & 56.5* & 72.2*\\
             GraphQA
            &\cite{baldassarre2021graphqa}  & \textit{Bioinform}&\textit{2021}
             &0.509 & 0.308& 0.329& 0.413&23.7* & 32.5* & 84.4* & 46.9* & 60.8*\\
             New IEConv
            &\cite{anonymous2022contrastive} & \textit{ICLR}&\textit{2022}
             & 0.735& 0.374& 0.544& 0.444&47.6* & 70.2* & 99.2* & 72.3* &  {87.2*}\\
             {GearNet}&\cite{zhang2022protein} & \textit{ICLR}&\textit{2023}   &  {0.810}& {0.400}&  {0.581}&  {0.430}&48.3 & 70.3 & 99.5 & 72.7 & 85.3\\
             ProNet&\cite{wanglearning2023}  & \textit{ICLR}&\textit{2023}  &  -&  - &  - & - & {52.7} &  {70.3} &  {99.3} &  {74.1} & 86.4\\
             {CDConv}&\cite{fancontinuous}  & \textit{ICLR}&\textit{2023}  &  {0.820} &  {0.453} &  {0.654} & 0.479 & {56.7} &  {77.7} &  {99.6} &  {78.0} & 88.5\\
             \hline \hline
             \multicolumn{2}{c||}{Ours}  & \multicolumn{2}{c||}{-} &  {\textbf{0.866}} &  {\textbf{0.474}} &  {\textbf{0.675}} & {\textbf{0.483}} & {\textbf{63.1}} &  {\textbf{81.2}} &  {99.6} &  {\textbf{81.3}} & {\textbf{89.6}} \\

            \hline
        \end{tabular}}
    \vspace{-1em}
\end{table*}

\subsection{Implementation Details} \label{subsec:impl}
\textbf{Network Architecture.} 
We progressively nominate $N_T$ from $N$ amino acids of the protein by $T$ iterations (see Figure~\ref{fig:pipeline}). We empirically set $T\!=\!4$, suggesting the neural clustering framework consists of four iterations.
At $t$-th iteration, we stack $B\!=\!2$ CRE blocks to learn the representation of the selected $N_t$ amino acids. 
In this way, our method is also a hierarchical framework that downsamples amino acids as the iteration proceeds.
In order to ensure a large enough receptive field at late itera-
 tions,  we increase the value of the cluster radius $r$ applied in SCI step as the number of iterations increases.
Concretely, the radii for the four iterations are  $r$, $2r$, $3r$, and $4r$, respectively. We adopt skip connection~\cite{he2016deep} per CRE block to facilitate information flow and ease network training.
Inspired by~\cite{ingraham2019generative,hermosilla2022contrastive,fancontinuous}, we\ms adopt\ms rotation\ms invariance (detailed in the appendix).

\noindent
\textbf{Training Objective.}
We follow the common protocol in this field~\cite{hermosilla2022contrastive,zhang2022protein,wang2022lm,fancontinuous} to append a fully connected neural network as the classification head at the tail of network. Softmax activation and cross-entropy loss are used for single-label classification, while Sigmoid function and binary cross-entropy loss for multi-label classification.

\noindent
\textbf{Reproducibility.}
We implement our method using PyTorch-Geometric library. For all our experiments, the training and testing are conducted on a single Nvidia RTX 3090 GPU with 24~GB memory. 
\section{Experiments}

We evaluate our method on four tasks following previous studies~\cite{hermosilla2020intrinsic, zhang2022protein, fancontinuous}: enzyme commission (EC) number prediction (\S\!~\ref{subsec:EC}), gene ontology (GO) term prediction (\S\!~\ref{subsec:GOTP}), protein fold classification (\S\!~\ref{subsec:PFC}), and enzyme reaction classification (\S\!~\ref{subsec:ERC}). 
Then, in \S\!~\ref{subsec:analysis}, we present a series of diagnostic studies. Finally, we provide a set of visual results in \S\!~\ref{sec:visualization} for in-depth analysis. 

\vspace{-3pt}
\subsection{Experiment on EC Number Prediction}\label{subsec:EC}

\vspace{-3pt}
\textbf{Task and Dataset.}  Enzyme Commission (EC) number prediction seeks to anticipate the EC numbers of diverse proteins that elucidate their role in catalyzing biochemical reactions. The EC numbers are chosen from the third and fourth levels of the EC tree, resulting in $538$ distinct binary classification tasks.
As in~\cite{gligorijevic2021structure}, the dataset of this task consists of $15,550$/$1,729$/$1,919$ proteins in \texttt{train}/\texttt{val}/\texttt{test} set, respectively.
For GO term and EC number prediction, we follow the multi-cutoff splits in~\cite{gligorijevic2021structure}
to ensure that the \texttt{test} set only contains PDB chains with a sequence identity of no more than 95\% to the proteins in the \texttt{train} set. 

\begin{table*}[!t]
    \footnotesize  
   \begin{minipage}[t]{0.66\textwidth}
   \caption{Ablative experiments for the neural clustering algorithm. (a) An off-the-shelf clustering algorithm; (b) A simple average pooling method; (c)   Randomly generate attention score $\gamma_k^n$. See \S\ref{subsec:analysis} for details.}
   \centering
    \resizebox{\textwidth}{!}{
		\setlength\tabcolsep{6pt}
		\renewcommand\arraystretch{1}
		\footnotesize
		\label{tab:ablation}
        \begin{tabular}{r|c|ccc|cccc|c}
            \thickhline 
            \rowcolor{mygray}\multirow{2}{*}{ } &
         
            \multirow{2}{*}{ }&
            \multicolumn{3}{c|}{ {GO}}&
            \multicolumn{4}{c|}{ {Fold Classification}}&
            \multirow{2}{*}{ }\\
            \cline{3-5}
            \cline{6-9}
            \rowcolor{mygray}\multirow{-2}{*}{Method}  & \multirow{-2}{*}{EC} & {BP} &  {MF}&  {CC} &  {Fold} &  {Super.} &  {Fam.} &  {Avg.} & \multirow{-2}{*}{Reaction}\\
            \hline \hline

              (a) 
             & 0.792 & 0.385 & 0.579 & 0.429& 43.1 & 67.1 & 99.1 & 69.8 & 86.8\\

              (b)                & 0.817& 0.452& 0.641&0.453 & 57.2 & 78.7 & 99.3 & 78.4& 88.1\\
              (c) 
             & 0.765& 0.342& 0.567& 0.415&44.6 & 69.5 & 99.2 & 71.1 & 86.4\\

             \hline \hline
             {Ours}  &  {\textbf{0.866}} &  {\textbf{0.474}} &  {\textbf{0.675}} & {\textbf{0.483}} & {\textbf{63.1}} &  {\textbf{81.2}} &  \textbf{99.6} &  {\textbf{81.3}} & {\textbf{89.6}} \\

            \hline
        \end{tabular}}
    \end{minipage}
    \hspace{0.1em}
    \begin{minipage}[t]{0.32\textwidth}
    \caption{Efficiency comparison to SOTA competitors on enzyme reaction classification. See \S\ref{subsec:analysis} for details}
        \label{tab:efficiency}
        \centering
        \resizebox{\textwidth}{!}{
		\setlength\tabcolsep{6pt}
		\renewcommand\arraystretch{1}
		\footnotesize
            \begin{tabular}{r|cc}
            \thickhline
            \rowcolor{mygray}{Method} & {Acc.$_{\!}$} & {$_{\!}$Runing$_{\!}$ Time}\\
            \hline \hline
            New\ms IEConv\ms \cite{anonymous2022contrastive} & 87.2\%$_{\!}$ & $_{\!}$75.3$_{\!}$ ms \\
            GearNet\ms \cite{zhang2022protein} & 85.3\%$_{\!}$ & $_{\!}$\textit{OOM}$_{\!}$ \\
            ProNet\ms \cite{wanglearning2023} & 86.4\%$_{\!}$ & $_{\!}$27.5$_{\!}$ ms \\
            CDConV\ms \cite{fancontinuous} & 88.5\%$_{\!}$ & $_{\!}$10.5$_{\!}$ ms \\ \hline \hline
            Ours & 89.6\%$_{\!}$ & $_{\!}$10.9$_{\!}$ ms \\
                \hline
            \end{tabular}}
    \end{minipage}
    \vspace{-10pt}
\end{table*}

\noindent
\textbf{Training Setup and Evaluation Metric.}
EC number prediction can be regarded as a multi-label classification task. The performance is evaluated by the protein-centric maximum F-score \textit{F}\textsubscript{max}, which is based on the precision and recall of the predictions for each protein.

\noindent
\textbf{Performance Comparison.}
We compare our neural clustering method with $11$ top-leading methods in Table~\ref{tab:all_result}. As seen, our method establishes a new state-of-the-art on EC number prediction task. It surpasses CDConv~\cite{fancontinuous} by \textbf{5.6}\% (0.820$\rightarrow$\textbf{0.866}) and GearNet~\cite{zhang2022protein} by \textbf{6.9}\% (0.810$\rightarrow$\textbf{0.866}) in terms of \textit{F}\textsubscript{max}. This indicates that our method can learn informative representations of proteins that reflect their functional roles in catalyzing biochemical reactions.


\vspace{-3pt}
\subsection{Experiment on GO Term Prediction}\label{subsec:GOTP}
\vspace{-3pt}
\textbf{Task and Dataset.} GO\ms term\ms prediction\ms aims\ms to\ms forecast\ms whether\ms a\ms protein\ms belongs\ms to\ms certain\ms GO\ms terms.\ms These\ms terms\ms categorize\ms proteins\ms into\ms functional\ms classes\ms that\ms are\ms hierarchically\ms related\ms and\ms organized\ms into\ms three\ms sub-tasks~\cite{gligorijevic2021structure}: molecular\ms function\ms (MF)\ms term\ms prediction\ms consisting\ms of $489$\ms classes,\ms biological\ms process\ms (BP)\ms term\ms prediction\ms including\ms $1,943$\ms classes,\ms cellular\ms component\ms (CC)\ms term\ms prediction\ms with\ms $320$\ms classes.\ms 
The\ms dataset\ms contains\ms $29,898$/$3,322$/$3,415$\ms proteins\ms for\ms \texttt{train}/\texttt{val}/\texttt{test},\ms respectively.

\noindent
\textbf{Training Setup and Evaluation Metric.}
GO term prediction is also a multi-label classification task. The protein-centric maximum F-score \textit{F}\textsubscript{max} is reported. 

\noindent
\textbf{Performance Comparison.}
We compare our method with $12$ existing state-of-the-art methods for protein representation learning on the task of predicting the GO term of proteins, where most of them are CNN or GNN-based methods. 
The results are shown in Table~\ref{tab:all_result}, where our framework achieves competitive \textit{F}\textsubscript{max} scores on all three sub-tasks, especially on MF (\textbf{0.675} \textit{vs} 0.654) and BP (\textbf{0.474} \textit{vs} 0.453) terms, compared to CDConv~\cite{fancontinuous}. Also, our method is clearly ahead of the second-best method, GearNet~\cite{zhang2022protein}, by large margins, \ie, \textbf{0.474} \textit{vs} 0.400 BP, \textbf{0.675} \textit{vs} 0.581 MF, \textbf{0.483} \textit{vs} 0.430 CC.
 Our new records across three sub-tasks show that our neural clustering method can learn rich representations of proteins that capture their functional diversity.        
\vspace{-15pt}
\subsection{Experiment on Protein Fold Classification}\label{subsec:PFC}
\vspace{-3pt}
\textbf{Task and Dataset.} Protein fold classification, firstly introduced in~\cite{hou2018deepsf}, aims to predict the fold class label of a protein. It contains three different evaluation scenarios: 1) Fold, where proteins~be- longing to the same superfamily are excluded during training, $12,312$/$736$/$718$ proteins for train/val/ test, 2) Superfamily, where proteins from the same family are not included during training, $12,312$/$736$/$1,254$ proteins for \texttt{train}/\texttt{val}/\texttt{test}, 3) Family, where proteins from the same family are used during training, $12,312$/$736$/$1,272$ proteins for \texttt{train}/\texttt{val}/\texttt{test}.

\noindent
\textbf{Training Setup and Evaluation Metric.}
Protein fold classification can be seen as a single-label classification task. Mean accuracy is used as the evaluation metric. 

\noindent
\textbf{Performance Comparison.}
In Table~\ref{tab:all_result}, we continue to compare our framework with these state-of-the-art methods on the task of classifying proteins into different fold classes. The fold class describes the overall shape and topology of a protein.
Our framework yields superior performance.
For example, it yields superior results as compared to  CDConv~\cite{fancontinuous} by \textbf{6.4}\%, ProNet~\cite{wang2022lm} by \textbf{10.4}\% and GearNet~\cite{zhang2022protein} by \textbf{14.8}\% on the Fold evaluation scenario. Considering that protein fold classification is challenging, such improvements are particularly impressive.
Across the board, our neural clustering framework surpasses all other methods of protein fold classification, demonstrating that our framework can learn robust representations of proteins that reflect their structural similarity.

\vspace{-3pt}
\subsection{Experiment on Enzyme Reaction Classification}\label{subsec:ERC}
 
\vspace{-3pt}
\textbf{Task and Dataset.}\ms Enzyme\ms reaction\ms classification\ms endeavors\ms to\ms predict\ms the\ms enzyme-catalyzed\ms reaction\ms class\ms of\ms a\ms protein,\ms utilizing\ms all\ms four\ms levels\ms of\ms the\ms EC\ms number\ms to\ms portray\ms reaction\ms class.\ms We\ms utilize\ms the\ms dataset\ms processed\ms by\ms \cite{hermosilla2020intrinsic}, which\ms consists\ms of\ms $384$ four-level\ms EC\ms classes\ms and $29,215$/$2,562$/$5,651$\ms proteins\ms for\ms \texttt{train}/\texttt{val}/\texttt{test},\ms where\ms proteins\ms have\ms less\ms than\ms 50\% sequence\ms similarity\ms in-between\ms splits.

\noindent
\textbf{Training Setup and Evaluation Metric.}
Enzyme reaction classification is regarded as a single-label classification task. We adopt Mean accuracy as the evaluation metric. 

\noindent
\textbf{Performance Comparison.}
Table~\ref{tab:all_result} presents comparison results of classifying proteins into different enzyme reaction classes. 
In terms of classification accuracy, our neural clustering framework outperforms the classic GCN-based method by a margin, \eg, GCN~\cite{kipf2017semi} by \textbf{22.3}\%, GAT~\cite{velickovic2018graph} by \textbf{34}\%, and GrahQA~\cite{baldassarre2021graphqa} by \textbf{28.8}\%. In addition, it surpasses recent three competitors, \ie, CDConv~\cite{fancontinuous} (\textbf{+1.1}\%), ProNet~\cite{wang2022lm} (\textbf{+3.2}\%), and GearNet~\cite{zhang2022protein} (\textbf{+4.3}\%). In summary, the proposed neural clustering framework achieves outstanding performance against state-of-the-art methods, suggesting that our method learns informative representations of proteins that reflect their catalytic activity.






\vspace{-3pt}
\subsection{Diagnose Analysis}\label{subsec:analysis}
\vspace{-3pt}

\textbf{Neural Clustering.}
To demonstrate the effectiveness of neural clustering, we compare it against three baseline approaches that employ naive methods as replacements.
Firstly, we use an off-the-shelf clustering algorithm, GRACLUS~\cite{dhillon2007weighted}, as a baseline (a). Secondly, we replace it with a simple average pooling method used in CDConv~\cite{fancontinuous} as a baseline (b). Lastly, we replace the attention score $\gamma_k^n$ with a random value as a baseline (c).
As shown in Table~\ref{tab:ablation}, our method significantly outperforms all three baselines. Specifically, it surpasses baseline (a) by an impressive \textbf{11.5}\%, baseline (b) by \textbf{2.5}\%, and baseline (c) by \textbf{10.2}\%.
The superior performance compared to baseline (a) highlights the importance of using a learnable clustering approach for effective representation learning. This demonstrates that our neural clustering is able to capture meaningful patterns and structures in the protein that are not captured by the off-the-shelf clustering algorithm.
Furthermore, comparison with baseline (c) supports the notion that learned assignment is more effective than random assignment, suggesting that\ms neural clustering can leverage the inherent structure and relationships of the protein to\ms make\ms informed\ms assignments,\ms leading\ms to\ms improved\ms performance.

\begin{table*}[t]
    \footnotesize  
  \begin{minipage}[t]{0.70\textwidth}
  \centering
  \caption{Analysis of a different number of iterations. See details in \S\ref{subsec:analysis}.}
    \label{tab:aba_roi}
    \resizebox{\textwidth}{!}{
		\setlength\tabcolsep{8pt}
		\renewcommand\arraystretch{1}
        \begin{tabular}{c||c|ccc|cccc|c}
            \thickhline 
            \rowcolor{mygray} &
       
            \multirow{2}{*}{ }&
            \multicolumn{3}{c|}{ {GO}}& 
            \multicolumn{4}{c|}{ {Fold Classification}}&
            \multirow{2}{*}{ }\\
            \cline{3-5}
            \cline{6-9}
            \rowcolor{mygray} \multirow{-2}{*}{$T$} & \multirow{-2}{*}{EC} & {BP} &  {MF}&  {CC} &   {Fold} &  {Super.} &  {Fam.} &  {Avg.} & \multirow{-2}{*}{Reaction}\\
            \hline \hline

   \textit{1} 
             & 0.717 & 0.402 &  0.593 & 0.432 &  55.7 & 73.2  & 97.4  & 75.4  & 84.7 \\
    \textit{2} 
             & 0.824 & 0.438 &  0.642 & 0.453 &  60.0 & 79.2  & 98.0 & 79.1  & 88.1 \\
    \textit{3} 
             & 0.855 & 0.469 &  0.677 & 0.480 &  62.2 & 80.8  & 99.3  & 80.8  & 89.0 \\
   \textit{4}
             & \textbf{0.866} & \textbf{0.474} & \textbf{0.675} & \textbf{0.483} & \textbf{63.1} &  \textbf{81.2}  &  \textbf{99.6} &  \textbf{81.3} & \textbf{89.6}   \\            
    \textit{5} 
             & 0.809 & 0.423 & 0.605 & 0.455 & 58.1  &  75.7  & 98.5  & 77.4  & 86.3 \\ 
 
            \hline
        \end{tabular}}      
    \end{minipage}
    \hspace{0.1em}
    \begin{minipage}[t]{0.29\textwidth}
 \centering
    \caption{$u\%$\ms\ms missing\ms\ms coordinates\ms\ms (\S\ref{subsec:analysis}).}
        \centering
        \resizebox{\textwidth}{!}{
		\setlength\tabcolsep{8pt}
		\renewcommand\arraystretch{1}

            \begin{tabular}{c||ccc}
            \thickhline
            \rowcolor{mygray}{ $u\%$ }  & {Fold} & {Super.}& {Fam.}\\
            \hline \hline
            {0\%} & \textbf{63.1} & \textbf{81.2}  & \textbf{99.6} \\
            {5\%} & { 61.9} &  79.8  & 99.5    \\
            {10\%} & { 60.1} & { 78.7}  & {99.5}    \\
            {20\%}  & { 56.7} &  76.9   & {99.3}  \\
            {30\%} & { 50.2} &  73.6  &  99.2   \\
            {40\%} & { 47.8} &  71.3  &  99.0   \\
                \hline
            \end{tabular}}
    \label{tab:ablation_missingcoor}\end{minipage}
    \vspace{-5pt}
\end{table*}

\begin{table*}[!h]
    \centering
    \vspace{-5pt}
    \caption{Comparison results with existing protein language models. See details in \S\ref{subsec:analysis}.}
    \label{tab:vs_pretrain}
    \resizebox{\textwidth}{!}{
		\setlength\tabcolsep{6pt}
		\renewcommand\arraystretch{1}
		\footnotesize
        \begin{tabular}{rl||rl||c|ccc|cccc|c}
            \thickhline 
            \rowcolor{mygray}\multicolumn{2}{c||}{\multirow{2}{*}{ }} &
            \multicolumn{2}{c||}{\multirow{2}{*}{ }} & 
            \multirow{2}{*}{ }&
            \multicolumn{3}{c|}{ {GO}}&
            \multicolumn{4}{c|}{ {Fold Classification}} & \\
            \cline{6-8}
            \cline{9-12}
            \rowcolor{mygray}\multirow{-2}{*}{Method}&& \multicolumn{2}{c||}{\multirow{-2}{*}{Pretraining Dataset}} & \multirow{-2}{*}{EC} & {BP} &  {MF}&  {CC} &  {Fold} &  {Super.} &  {Fam.} &  {Avg.} & \multirow{-2}{*}{Reaction}\\
            \hline \hline
             DeepFRI & \cite{ozturk2018deepdta} & Pfam & 10M
             & 0.631 & 0.399& 0.465& 0.460& 15.3 &  20.6 &  73.2 & 36.4 & 63.3 \\
             ESM-1b
            &\cite{rives2021biological} & UniRef50 & 24M
            & 0.864 & 0.470 &0.657& 0.488
            & 26.8 &  60.1 &  97.8 & 61.6 & 83.1 \\
            ProtBERT-BFD
            &\cite{elnaggar2021prottrans} & BFD & 2.1B
            & 0.838 & 0.279 & 0.456 & 0.408
            & 26.6 & 55.8 & 97.6 & 60.0 & 72.2 \\  
            IEConv (amino level)
            &\cite{hermosilla2022contrastive} & PDB & 476K
            & - & 0.468 & 0.661 & 0.516
            & 50.3 & 80.6 & 99.7 & 76.9 & 88.1 \\   
            LM-GVP
            &\cite{wang2022lm} & UniRef100 & 0.21B
            & 0.664 & 0.417 &0.545 &0.527
            & - & - & - & - & - \\   
            GearNet-Edge-IEConv
            &\cite{zhang2022protein} & AlphaFoldDB & 805K
            & 0.874 & 0.490 &0.654 &0.488
            & 54.1 &80.5 &99.9 & 78.2 & 87.5 \\        
             \hline \hline
             Ours &  & \multicolumn{2}{c||}{-} &  {{0.866}} &  {{0.474}} &  {\textbf{0.675}} & {{0.483}} & {\textbf{63.1}} &  {\textbf{81.2}} &  {99.6} &  {\textbf{81.3}} & \textbf{89.6}\\

            \hline
        \end{tabular}}
        \vspace{-5pt}
\end{table*}

\begin{figure*}[h!]
    \centering
    \begin{minipage}[t]{0.25\textwidth}
         \vspace{12pt}
       \centering
      \setlength{\abovecaptionskip}{0.3cm}
      \includegraphics[width=\linewidth]{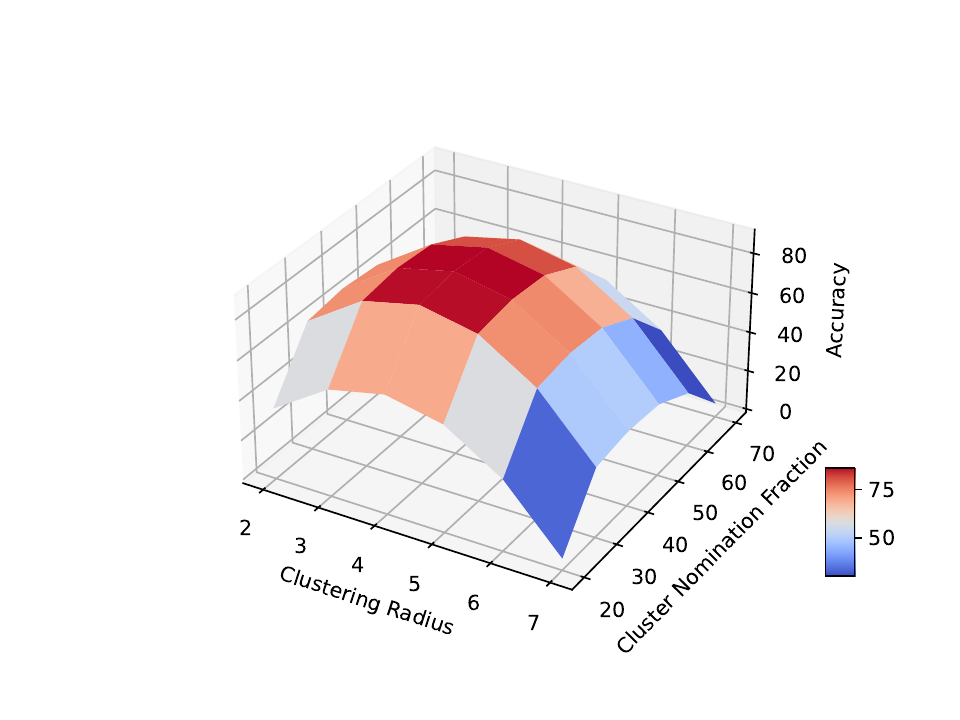}
      \caption{Performance change curve with different combinations of $\omega$ and $r$ for enzyme reaction classification. See \S\!~\ref{subsec:analysis} for details.}
\label{fig:performance_analysis}	
     \end{minipage}
    \hspace{0.1em}
    \begin{minipage}[t]{0.72\textwidth}
    \vspace{0pt}
    \centering
    \includegraphics[width=\textwidth]{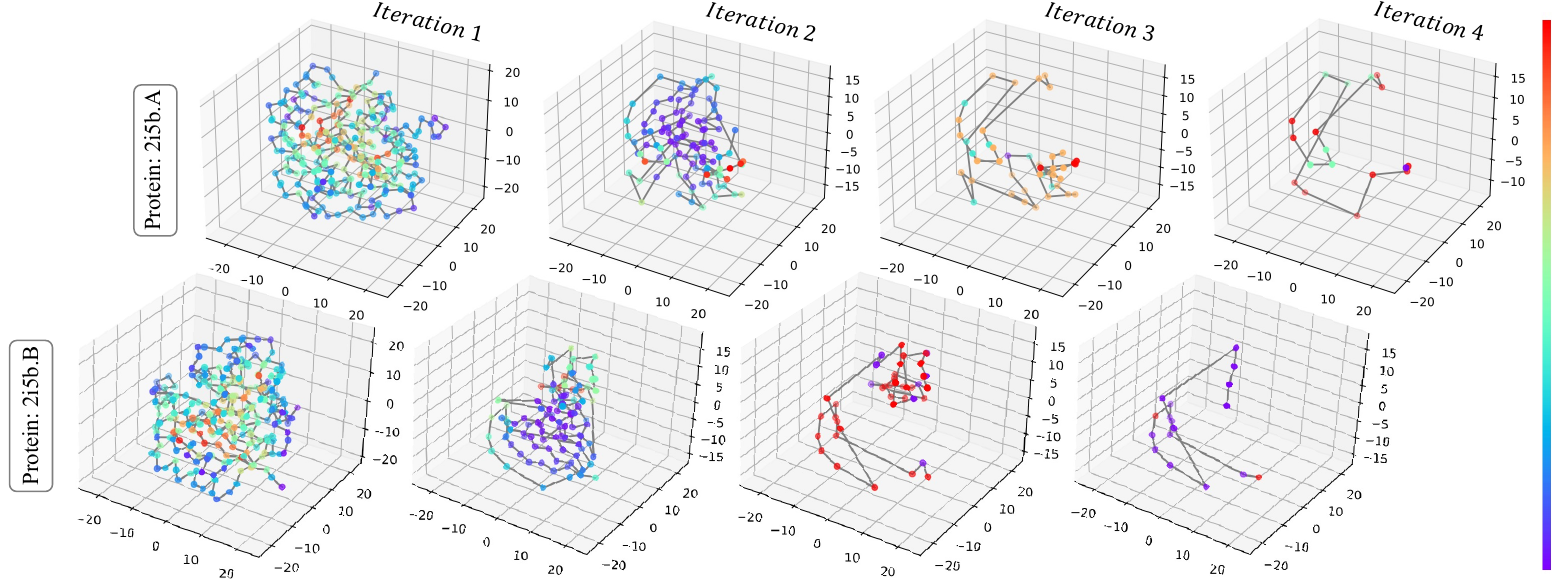}
    \vspace{-15pt}
    \caption{Visualization results of the protein structure at each iteration. The color of the node denotes the score calculated in CN step. See related analysis in~\S\!~\ref{sec:visualization}.}
        \label{fig:visualization}
     \end{minipage} 
     \vspace{-15pt}
\end{figure*}

\begin{figure*}[t]
    \centering
    \vspace{-3pt}
    \includegraphics[width=0.96\textwidth]{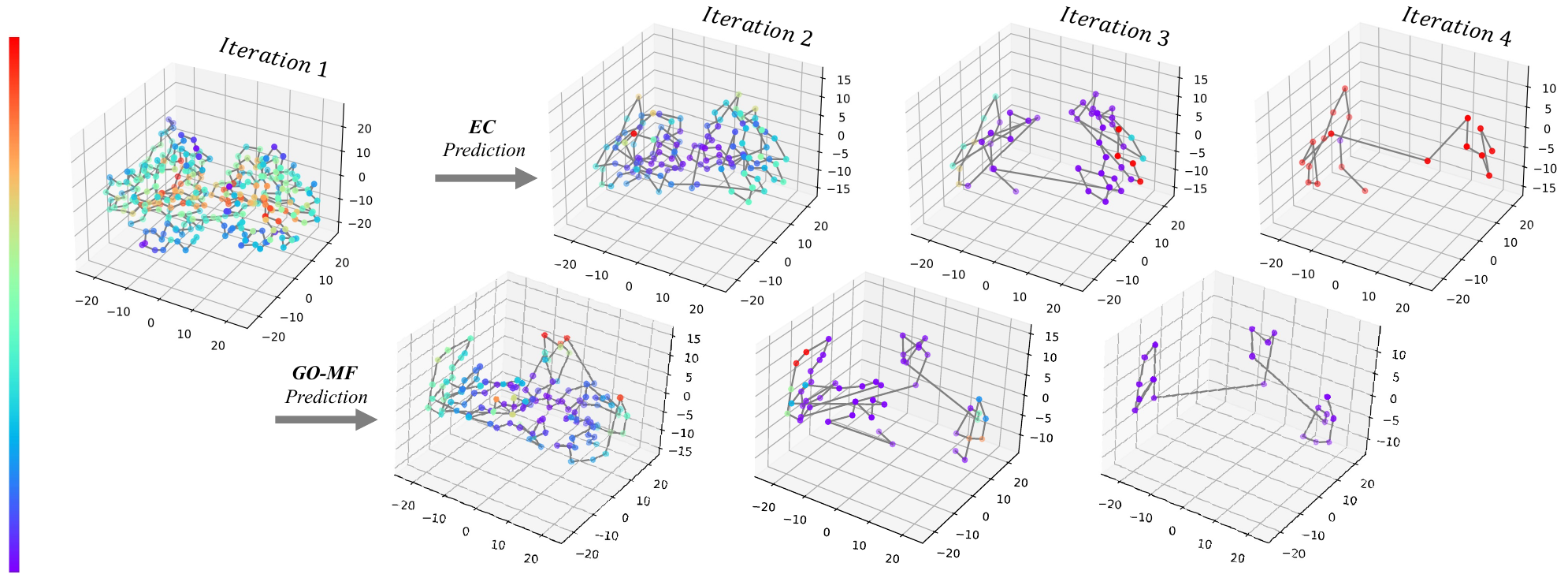}
    \vspace{-8pt}
    \caption{Clustering$_{\!}$ results$_{\!}$ for$_{\!}$ a$_{\!}$ protein$_{\!}$ exhibit$_{\!}$ variations$_{\!}$ across$_{\!}$ EC$_{\!}$ and$_{\!}$ GO-MF$_{\!}$ predictions$_{\!}$. See related analysis in~\S\!~\ref{sec:visualization}.}
    \vspace{-6pt}
    \label{fig:sameproteinvis}
\end{figure*}

\begin{figure*}[t]
    \centering
    \includegraphics[width=0.95\textwidth]{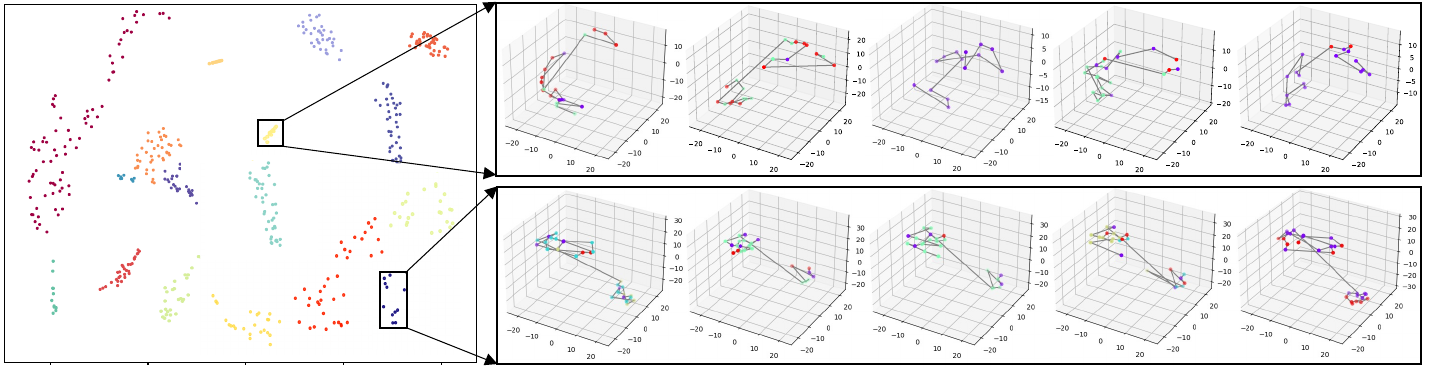}
    \vspace{-10pt}
    \caption{UMAP projection~\cite{mcinnes2018umap} of the learned representation. See related analysis in~\S\!~\ref{sec:visualization}.}
    \vspace{-15pt}
    \label{vis:umap}
\end{figure*}

\noindent
\textbf{Efficiency.} 
We conduct an investigation into the efficiency of our neural clustering-based framework, focusing on its running time for enzyme reaction classification. The mean running time per prediction was measured using a single Nvidia RTX $3090$ GPU, and the results are summarized in Table~\ref{tab:efficiency}. In particular, GearNet~\cite{zhang2022protein}, a competing method known for its high complexity, cannot be trained using the same GPU due to its resource requirements.
Notably, our method achieves state-of-the-art performance while maintaining a comparable running time to existing approaches, suggesting the efficiency of our method.

\noindent
\textbf{Initial Clustering Radius.}
The initial clustering radius $r$ determines the cluster size formed in SCI step.
A larger radius leads to more amino acids in each cluster, potentially capturing a greater amount of spatial structural information. However, this also introduces the risk of increased noise within the clusters. Conversely, a smaller radius results in fewer amino acids being included in each cluster, which can reduce noise but may also lead to the loss of some critical information.
Therefore, we conducted experiments ranging from $2.0$ to $7.0$ and assessed the performance on protein fold classification and enzyme reaction classification. The experimental results, presented in Figure~\ref{fig:performance_analysis},
indicate that the optimal performance is achieved when $r\!=\!4.0$, suggesting a suitable balance between capturing sufficient structural information and mitigating the detrimental effects of noise.

\noindent
\textbf{Cluster Nomination Fraction.}
In CN step, a cluster nomination fraction $\omega$ determines the proportion of clusters 
selected as the medoid nodes for the next 
iteration. A larger $\omega$ means that more clusters are retained, which may preserve more information but also increase redundancy and complexity. While a smaller $\omega$ means that fewer clusters are retained, which may reduce redundancy and complexity but also lose some information. We experiment with different values of $\omega$ from $20\%$ to $70\%$ and report results on protein fold classification and enzyme reaction classification. As shown in Figure~\ref{fig:performance_analysis},  
the best performance is achieved when $\omega\!=\!40\%$, suggesting a suitable trade-off between preserving information and reducing redundancy and complexity.

\noindent
\textbf{Number of Iterations.} Table~\ref{tab:aba_roi} also studies the impact of the number of iterations. For enzyme reaction classification, increasing $T$ from 1 to 4 leads to better performance (\ie, 84.7\%$\rightarrow$89.6\%). However, the accuracy drops significantly from 89.6\% to 86.3\% when $T$ is set as $5$. This may be because over-clustering finds insufficient and insignificant amino acids, which are harmful to representation learning. Similar trends can be observed in the results of other tasks.

\noindent
\textbf{Percentages of Missing Coordinates.}
In some cases, the protein structures may have missing coordinates due to experimental errors or incomplete data. To test the robustness of our framework to handle such cases, we randomly remove a certain percentage $u\%$ of coordinates from the protein structures and evaluate our framework on protein fold classification. The results are shown in Table~\ref{tab:ablation_missingcoor}, where we can find that our framework still achieves competitive performance when some of the coordinates are missing. For instance, in Superfamily evaluation scenario, our framework achieves an average accuracy of 78.7\% when 10\% of the coordinates are missing, which is only slightly lower than the accuracy of 81.2\% when no coordinates are missing. This indicates that our framework learns robust representations of proteins that are not sensitive to missing coordinates.

\noindent
\textbf{Comparison to Existing Protein Language Models.}
\revise{To further showcase the effectiveness of our method, we compare our algorithm with some recent protein pretraining language models on fold classification~\cite{rives2021biological,elnaggar2021prottrans,hermosilla2022contrastive,wang2022lm,zhang2022protein}, which are typically pre-trained with a large scale of data. 
As shown in Table~\ref{tab:vs_pretrain}, our method still yields better results without any pre-training or self-supervised learning. It may be that one or a few self-supervised tasks are insufficient to learn effective representations, as mentioned in~\cite{fancontinuous}.} This sheds light on the direction of our future efforts: it is interesting to embrace our algorithm with existing protein language models since our core idea is principled.

\subsection{Visualization}\label{sec:visualization}

We visualize the protein structure at each iteration in Figure~\ref{fig:visualization}. 
The color of the node corresponds to the score calculated in CN step. 
By using such an iterative clustering algorithm, this method is supposed to explore the critical amino acids of the protein.
For example, we examine the protein `2i5b.A', characterized by a complex structure consisting of numerous loops and helices. 
After the first iteration of clustering, our method selects some amino acids that are located at the ends or bends of these loops and helices as the medoid nodes for the next iteration.  
Subsequently, after the second clustering iteration, our method further narrows down the number of amino acids by selecting those that have high scores. 
Ultimately, our method identifies a small subset of amino acids with the highest scores, which are seen as the representative ones for the protein.
When comparing the visualization results of protein chain pairs stemming from the same family or protein, \eg, `2i5b.A' \textit{vs} `2i5b.B', we observe remarkable similarity in their clustering outcomes, suggesting that they possess critical amino acids fundamentally determining their respective structures and functionalities. This further validates that our method is effective in identifying these critical amino acids.

In Figure~\ref{fig:sameproteinvis}, we present the clustering results of the same protein for different tasks: EC and GO-MF. Interestingly, we observe variations in the results for the same protein across different tasks, indicating that different critical components are identified for different functions.
Moreover, some parts are highlighted for both two tasks. This is probably because these parts are informative across tasks.
To further prove that our method can indeed discover some functional motifs of proteins, following LM-GV~\cite{wang2022lm}, we apply UMAP~\cite{mcinnes2018umap} to analyze the learned representation at the penultimate layer on Enzyme reaction classification and use DBSCAN32~\cite{ester1996density} to extract protein clusters. $20$ of $384$ classes are shown in Figure~\ref{vis:umap}, where two clusters are selected for detailed analysis. It is clear that proteins originating from the same cluster exhibit similar structural motifs, as generated by our method. This compelling observation underscores the efficacy of our clustering approach in identifying proteins possessing analogous structural motifs that are related to their enzyme reaction functions.

\section{Conclusion}
In this work, our epistemology is centered on protein representation learning by a neural clustering paradigm, which coins a compact and powerful framework to unify the community of protein science and respect the distinctive characteristics of each sub-task. The clustering insight leads us to introduce new approaches for spherical cluster initialization, cluster representation extraction, and cluster nomination based on both 1D and 3D information of amino acids. Empirical results suggest that our framework achieves superior performance in all four sub-tasks. Our research may potentially benefit the broader domain of bioinformatics and computational biology as a whole.

\clearpage
{
    \small
    \bibliographystyle{ieeenat_fullname}
    \bibliography{main}

\begin{thebibliography}{95}
\providecommand{\natexlab}[1]{#1}
\providecommand{\url}[1]{\texttt{#1}}
\expandafter\ifx\csname urlstyle\endcsname\relax
  \providecommand{\doi}[1]{doi: #1}\else
  \providecommand{\doi}{doi: \begingroup \urlstyle{rm}\Url}\fi

\bibitem[Abualigah et~al.(2022)Abualigah, Almotairi, Al-qaness, Ewees, Yousri, Abd~Elaziz, and Nadimi-Shahraki]{abualigah2022efficient}
Laith Abualigah, Khaled~H Almotairi, Mohammed~AA Al-qaness, Ahmed~A Ewees, Dalia Yousri, Mohamed Abd~Elaziz, and Mohammad~H Nadimi-Shahraki.
\newblock Efficient text document clustering approach using multi-search arithmetic optimization algorithm.
\newblock \emph{Knowledge-Based Systems}, 248:\penalty0 108833, 2022.

\bibitem[Aggarwal and Zhai(2012)]{aggarwal2012survey}
Charu~C Aggarwal and ChengXiang Zhai.
\newblock A survey of text clustering algorithms.
\newblock \emph{Mining text data}, pages 77--128, 2012.

\bibitem[Ali et~al.(2023)Ali, Vascon, Stadelmann, and Pelillo]{ali2023quasi}
Waqar Ali, Sebastiano Vascon, Thilo Stadelmann, and Marcello Pelillo.
\newblock Quasi-cliquepool: Hierarchical graph pooling for graph classification.
\newblock In \emph{SAC}, 2023.

\bibitem[Amidi et~al.(2018)Amidi, Amidi, Vlachakis, Megalooikonomou, Paragios, and Zacharaki]{amidi2018enzynet}
Afshine Amidi, Shervine Amidi, Dimitrios Vlachakis, Vasileios Megalooikonomou, Nikos Paragios, and Evangelia~I Zacharaki.
\newblock Enzynet: enzyme classification using 3d convolutional neural networks on spatial representation.
\newblock \emph{PeerJ}, 6:\penalty0 e4750, 2018.

\bibitem[Asgari and Mofrad(2015)]{asgari2015continuous}
Ehsaneddin Asgari and Mohammad~RK Mofrad.
\newblock Continuous distributed representation of biological sequences for deep proteomics and genomics.
\newblock \emph{PloS ONE}, 10\penalty0 (11):\penalty0 e0141287, 2015.

\bibitem[Baek et~al.(2021)Baek, Kang, and Hwang]{baek2021accurate}
Jinheon Baek, Minki Kang, and Sung~Ju Hwang.
\newblock Accurate learning of graph representations with graph multiset pooling.
\newblock In \emph{ICLR}, 2021.

\bibitem[Baldassarre et~al.(2021)Baldassarre, Men{\'e}ndez~Hurtado, Elofsson, and Azizpour]{baldassarre2021graphqa}
Federico Baldassarre, David Men{\'e}ndez~Hurtado, Arne Elofsson, and Hossein Azizpour.
\newblock Graphqa: protein model quality assessment using graph convolutional networks.
\newblock \emph{Bioinformatics}, 37\penalty0 (3):\penalty0 360--366, 2021.

\bibitem[Blundell and Wood(1975)]{blundell1975evolution}
TL Blundell and SP Wood.
\newblock Is the evolution of insulin darwinian or due to selectively neutral mutation?
\newblock \emph{Nature}, 257\penalty0 (5523):\penalty0 197--203, 1975.

\bibitem[Branden and Tooze(2012)]{branden2012introduction}
Carl~Ivar Branden and John Tooze.
\newblock \emph{Introduction to protein structure}.
\newblock Garland Science, 2012.

\bibitem[Cao et~al.(2021)Cao, Das, Chenthamarakshan, Chen, Melnyk, and Shen]{cao2021fold2seq}
Yue Cao, Payel Das, Vijil Chenthamarakshan, Pin-Yu Chen, Igor Melnyk, and Yang Shen.
\newblock Fold2seq: A joint sequence (1d)-fold (3d) embedding-based generative model for protein design.
\newblock In \emph{ICML}, 2021.

\bibitem[Chen et~al.(2024)Chen, Li, Yang, and Wang]{guikun2024}
Guikun Chen, Xia Li, Yi Yang, and Wenguan Wang.
\newblock Neural clustering based visual representation learning.
\newblock In \emph{CVPR}, 2024.

\bibitem[Chen et~al.(2021)Chen, Liu, and Jia]{chen2021jigsaw}
Pengguang Chen, Shu Liu, and Jiaya Jia.
\newblock Jigsaw clustering for unsupervised visual representation learning.
\newblock In \emph{CVPR}, 2021.

\bibitem[Chou and Cai(2003)]{chou2003predicting}
Kuo-Chen Chou and Yu-Dong Cai.
\newblock Predicting protein quaternary structure by pseudo amino acid composition.
\newblock \emph{Proteins: Structure, Function, and Bioinformatics}, 53\penalty0 (2):\penalty0 282--289, 2003.

\bibitem[Close and Kashef(2020)]{close2020combining}
Liam Close and Rasha Kashef.
\newblock Combining artificial immune system and clustering analysis: A stock market anomaly detection model.
\newblock \emph{Journal of Intelligent Learning Systems and Applications}, 12\penalty0 (04):\penalty0 83--108, 2020.

\bibitem[Derevyanko et~al.(2018)Derevyanko, Grudinin, Bengio, and Lamoureux]{derevyanko2018deep}
Georgy Derevyanko, Sergei Grudinin, Yoshua Bengio, and Guillaume Lamoureux.
\newblock Deep convolutional networks for quality assessment of protein folds.
\newblock \emph{Bioinformatics}, 34\penalty0 (23):\penalty0 4046--4053, 2018.

\bibitem[Dhillon et~al.(2007)Dhillon, Guan, and Kulis]{dhillon2007weighted}
Inderjit~S Dhillon, Yuqiang Guan, and Brian Kulis.
\newblock Weighted graph cuts without eigenvectors a multilevel approach.
\newblock \emph{IEEE TPAMI}, 29\penalty0 (11):\penalty0 1944--1957, 2007.

\bibitem[Ding et~al.(2024)Ding, Li, Wang, and Yang]{ding2024s2vnet}
Yuhang Ding, Liulei Li, Wenguan Wang, and Yi Yang.
\newblock Clustering propagation for universal medical image segmentation.
\newblock In \emph{CVPR}, 2024.

\bibitem[Doolittle(1981)]{doolittle1981similar}
Russell~F Doolittle.
\newblock Similar amino acid sequences: chance or common ancestry?
\newblock \emph{Science}, 214\penalty0 (4517):\penalty0 149--159, 1981.

\bibitem[Duvenaud et~al.(2015)Duvenaud, Maclaurin, Iparraguirre, Bombarell, Hirzel, Aspuru-Guzik, and Adams]{duvenaud2015convolutional}
David~K Duvenaud, Dougal Maclaurin, Jorge Iparraguirre, Rafael Bombarell, Timothy Hirzel, Al{\'a}n Aspuru-Guzik, and Ryan~P Adams.
\newblock Convolutional networks on graphs for learning molecular fingerprints.
\newblock In \emph{NeurIPS}, 2015.

\bibitem[Elnaggar et~al.(2021)Elnaggar, Heinzinger, Dallago, Rehawi, Wang, Jones, Gibbs, Feher, Angerer, Steinegger, et~al.]{elnaggar2021prottrans}
Ahmed Elnaggar, Michael Heinzinger, Christian Dallago, Ghalia Rehawi, Yu Wang, Llion Jones, Tom Gibbs, Tamas Feher, Christoph Angerer, Martin Steinegger, et~al.
\newblock Prottrans: Towards cracking the language of lifes code through self-supervised deep learning and high performance computing.
\newblock \emph{IEEE TPAMI}, 2021.

\bibitem[Ester et~al.(1996)Ester, Kriegel, Sander, Xu, et~al.]{ester1996density}
Martin Ester, Hans-Peter Kriegel, J{\"o}rg Sander, Xiaowei Xu, et~al.
\newblock A density-based algorithm for discovering clusters in large spatial databases with noise.
\newblock In \emph{KDD}, 1996.

\bibitem[Fan et~al.(2023{\natexlab{a}})Fan, Wang, Yang, and Kankanhalli]{fancontinuous}
Hehe Fan, Zhangyang Wang, Yi Yang, and Mohan Kankanhalli.
\newblock Continuous-discrete convolution for geometry-sequence modeling in proteins.
\newblock In \emph{ICLR}, 2023{\natexlab{a}}.

\bibitem[Fan et~al.(2023{\natexlab{b}})Fan, Zhu, Yang, and Kankanhalli]{Fan_2023_CVPR}
Hehe Fan, Linchao Zhu, Yi Yang, and Mohan Kankanhalli.
\newblock Pointlistnet: Deep learning on 3d point lists.
\newblock In \emph{CVPR}, 2023{\natexlab{b}}.

\bibitem[Feng et~al.(2023)Feng, Wang, Wang, Yang, and Zheng]{feng2023clustering}
Tuo Feng, Wenguan Wang, Xiaohan Wang, Yi Yang, and Qinghua Zheng.
\newblock Clustering based point cloud representation learning for 3d analysis.
\newblock In \emph{ICCV}, 2023.

\bibitem[Feng et~al.(2024)Feng, Quan, Wang, Wang, and Yang]{feng2024I3D}
Tuo Feng, Ruijie Quan, Xiaohan Wang, Wenguan Wang, and Yi Yang.
\newblock Interpretable3d: an ad-hoc interpretable classifier for 3d point clouds.
\newblock In \emph{AAAI}, 2024.

\bibitem[Frazier et~al.(1972)Frazier, Angeletti, and Bradshaw]{frazier1972nerve}
William~A Frazier, Ruth~Hogue Angeletti, and Ralph~A Bradshaw.
\newblock Nerve growth factor and insulin: Structural similarities indicate an evolutionary relationship reflected by physiological action.
\newblock \emph{Science}, 176\penalty0 (4034):\penalty0 482--488, 1972.

\bibitem[Gao and Ji(2019)]{gao2019graph}
Hongyang Gao and Shuiwang Ji.
\newblock Graph u-nets.
\newblock In \emph{ICML}, 2019.

\bibitem[Gao et~al.(2021{\natexlab{a}})Gao, Liu, and Ji]{gao2021topology}
Hongyang Gao, Yi Liu, and Shuiwang Ji.
\newblock Topology-aware graph pooling networks.
\newblock \emph{IEEE TPAMI}, 43\penalty0 (12):\penalty0 4512--4518, 2021{\natexlab{a}}.

\bibitem[Gao et~al.(2021{\natexlab{b}})Gao, Dai, Li, Xiong, and Frossard]{gao2021ipool}
Xing Gao, Wenrui Dai, Chenglin Li, Hongkai Xiong, and Pascal Frossard.
\newblock ipool-information-based pooling in hierarchical graph neural networks.
\newblock \emph{IEEE TNNLS}, 33\penalty0 (9):\penalty0 5032--5044, 2021{\natexlab{b}}.

\bibitem[Gligorijevi{\'c} et~al.(2021)Gligorijevi{\'c}, Renfrew, Kosciolek, Leman, Berenberg, Vatanen, Chandler, Taylor, Fisk, Vlamakis, et~al.]{gligorijevic2021structure}
Vladimir Gligorijevi{\'c}, P~Douglas Renfrew, Tomasz Kosciolek, Julia~Koehler Leman, Daniel Berenberg, Tommi Vatanen, Chris Chandler, Bryn~C Taylor, Ian~M Fisk, Hera Vlamakis, et~al.
\newblock Structure-based protein function prediction using graph convolutional networks.
\newblock \emph{Nature Communications}, 12\penalty0 (1):\penalty0 3168, 2021.

\bibitem[Govindarajan et~al.(1999)Govindarajan, Recabarren, and Goldstein]{govindarajan1999estimating}
Sridhar Govindarajan, Ruben Recabarren, and Richard~A Goldstein.
\newblock Estimating the total number of protein folds.
\newblock \emph{Proteins: Structure, Function, and Bioinformatics}, 35\penalty0 (4):\penalty0 408--414, 1999.

\bibitem[Govindasamy et~al.(2018)Govindasamy, Arumugam, Zhuang, Kelley, and Vellangany]{govindasamy2018cluster}
Ramu Govindasamy, Surendran Arumugam, Jingkun Zhuang, Kathleen~M Kelley, and Isaac Vellangany.
\newblock Cluster analysis of wine market segmentation-a consumer based study in the mid-atlantic usa.
\newblock \emph{Economic Affairs}, 63\penalty0 (1):\penalty0 151--157, 2018.

\bibitem[Hamilton et~al.(2017)Hamilton, Ying, and Leskovec]{hamilton2017inductive}
Will Hamilton, Zhitao Ying, and Jure Leskovec.
\newblock Inductive representation learning on large graphs.
\newblock In \emph{NeurIPS}, 2017.

\bibitem[Hartigan and Wong(1979)]{hartigan1979algorithm}
John~A Hartigan and Manchek~A Wong.
\newblock Algorithm as 136: A k-means clustering algorithm.
\newblock \emph{Journal of the royal statistical society. series c (applied statistics)}, 28\penalty0 (1):\penalty0 100--108, 1979.

\bibitem[He et~al.(2016)He, Zhang, Ren, and Sun]{he2016deep}
Kaiming He, Xiangyu Zhang, Shaoqing Ren, and Jian Sun.
\newblock Deep residual learning for image recognition.
\newblock In \emph{CVPR}, 2016.

\bibitem[Hermosilla and Ropinski(2022{\natexlab{a}})]{anonymous2022contrastive}
Pedro Hermosilla and Timo Ropinski.
\newblock Contrastive representation learning for 3d protein structures.
\newblock In \emph{ICLR}, 2022{\natexlab{a}}.

\bibitem[Hermosilla and Ropinski(2022{\natexlab{b}})]{hermosilla2022contrastive}
Pedro Hermosilla and Timo Ropinski.
\newblock Contrastive representation learning for 3d protein structures.
\newblock \emph{arXiv preprint arXiv:2205.15675}, 2022{\natexlab{b}}.

\bibitem[Hermosilla et~al.(2021)Hermosilla, Sch{\'a}fer, Lang, Fackelmann, V{\'a}zquez, Kozl{\'\i}kov{\'a}, Krone, Ritschel, and Ropinski]{hermosilla2020intrinsic}
Pedro Hermosilla, Marco Sch{\'a}fer, Mat{\v{e}}j Lang, Gloria Fackelmann, Pere~Pau V{\'a}zquez, Barbora Kozl{\'\i}kov{\'a}, Michael Krone, Tobias Ritschel, and Timo Ropinski.
\newblock Intrinsic-extrinsic convolution and pooling for learning on 3d protein structures.
\newblock In \emph{ICLR}, 2021.

\bibitem[Hou et~al.(2018)Hou, Adhikari, and Cheng]{hou2018deepsf}
Jie Hou, Badri Adhikari, and Jianlin Cheng.
\newblock Deepsf: deep convolutional neural network for mapping protein sequences to folds.
\newblock \emph{Bioinformatics}, 34\penalty0 (8):\penalty0 1295--1303, 2018.

\bibitem[Ingraham et~al.(2019)Ingraham, Garg, Barzilay, and Jaakkola]{ingraham2019generative}
John Ingraham, Vikas Garg, Regina Barzilay, and Tommi Jaakkola.
\newblock Generative models for graph-based protein design.
\newblock In \emph{NeurIPS}, 2019.

\bibitem[Ingram(2004)]{ingram2004sickle}
Vernon~M Ingram.
\newblock Sickle-cell anemia hemoglobin: the molecular biology of the first ``molecular disease''-the crucial importance of serendipity.
\newblock \emph{Genetics}, 167\penalty0 (1):\penalty0 1--7, 2004.

\bibitem[Jaiswal et~al.(2020)Jaiswal, Kaushal, Singh, and Biswas]{jaiswal2020green}
Deepak Jaiswal, Vikrant Kaushal, Pankaj~Kumar Singh, and Abhijeet Biswas.
\newblock Green market segmentation and consumer profiling: a cluster approach to an emerging consumer market.
\newblock \emph{Benchmarking: An International Journal}, 28\penalty0 (3):\penalty0 792--812, 2020.

\bibitem[Janani and Vijayarani(2019)]{janani2019text}
R Janani and S Vijayarani.
\newblock Text document clustering using spectral clustering algorithm with particle swarm optimization.
\newblock \emph{Expert Systems with Applications}, 134:\penalty0 192--200, 2019.

\bibitem[Jing et~al.(2021)Jing, Eismann, Soni, and Dror]{jing2021equivariant}
Bowen Jing, Stephan Eismann, Pratham~N. Soni, and Ron~O. Dror.
\newblock Learning from protein structure with geometric vector perceptrons.
\newblock In \emph{ICLR}, 2021.

\bibitem[Kipf and Welling(2017{\natexlab{a}})]{kipf2016semi}
Thomas~N Kipf and Max Welling.
\newblock Semi-supervised classification with graph convolutional networks.
\newblock In \emph{ICML}, 2017{\natexlab{a}}.

\bibitem[Kipf and Welling(2017{\natexlab{b}})]{kipf2017semi}
Thomas~N. Kipf and Max Welling.
\newblock Semi-supervised classification with graph convolutional networks.
\newblock In \emph{ICLR}, 2017{\natexlab{b}}.

\bibitem[Knyazev et~al.(2019)Knyazev, Taylor, and Amer]{knyazev2019understanding}
Boris Knyazev, Graham~W Taylor, and Mohamed Amer.
\newblock Understanding attention and generalization in graph neural networks.
\newblock In \emph{NeurIPS}, 2019.

\bibitem[Kulmanov and Hoehndorf(2020)]{kulmanov2020deepgoplus}
Maxat Kulmanov and Robert Hoehndorf.
\newblock Deepgoplus: improved protein function prediction from sequence.
\newblock \emph{Bioinformatics}, 36\penalty0 (2):\penalty0 422--429, 2020.

\bibitem[Kulmanov et~al.(2018)Kulmanov, Khan, and Hoehndorf]{kulmanov2018deepgo}
Maxat Kulmanov, Mohammed~Asif Khan, and Robert Hoehndorf.
\newblock Deepgo: predicting protein functions from sequence and interactions using a deep ontology-aware classifier.
\newblock \emph{Bioinformatics}, 34\penalty0 (4):\penalty0 660--668, 2018.

\bibitem[Lee et~al.(2019)Lee, Lee, and Kang]{lee2019self}
Junhyun Lee, Inyeop Lee, and Jaewoo Kang.
\newblock Self-attention graph pooling.
\newblock In \emph{ICML}, 2019.

\bibitem[Liang et~al.(2022)Liang, Wang, Miao, and Yang]{liang2022gmmseg}
Chen Liang, Wenguan Wang, Jiaxu Miao, and Yi Yang.
\newblock Gmmseg: Gaussian mixture based generative semantic segmentation models.
\newblock In \emph{NeurIPS}, 2022.

\bibitem[Liang et~al.(2023)Liang, Zhou, Liu, and Wang]{liang2023clustseg}
James Liang, Tianfei Zhou, Dongfang Liu, and Wenguan Wang.
\newblock Clustseg: Clustering for universal segmentation.
\newblock In \emph{ICML}, 2023.

\bibitem[Liu et~al.(2023)Liu, Zhan, Wu, Li, Du, Hu, Liu, and Tao]{liu2022graph}
Chuang Liu, Yibing Zhan, Jia Wu, Chang Li, Bo Du, Wenbin Hu, Tongliang Liu, and Dacheng Tao.
\newblock Graph pooling for graph neural networks: Progress, challenges, and opportunities.
\newblock In \emph{IJCAI}, 2023.

\bibitem[Lloyd(1982)]{lloyd1982least}
Stuart Lloyd.
\newblock Least squares quantization in pcm.
\newblock \emph{IEEE Transactions on Information Theory}, 28\penalty0 (2):\penalty0 129--137, 1982.

\bibitem[Lu et~al.(2024)Lu, Quan, Zhu, and Yang]{lu2024zero}
Yu Lu, Ruijie Quan, Linchao Zhu, and Yi Yang.
\newblock Zero-shot video grounding with pseudo query lookup and verification.
\newblock \emph{IEEE TIP}, 33:\penalty0 1643--1654, 2024.

\bibitem[Luo et~al.(2022)Luo, Ju, Qu, Gu, Chen, Deng, Hua, and Zhang]{luo2022clear}
Xiao Luo, Wei Ju, Meng Qu, Yiyang Gu, Chong Chen, Minghua Deng, Xian-Sheng Hua, and Ming Zhang.
\newblock Clear: Cluster-enhanced contrast for self-supervised graph representation learning.
\newblock \emph{IEEE TNNLS}, 2022.

\bibitem[Luzhnica et~al.(2019)Luzhnica, Day, and Lio]{luzhnica2019clique}
Enxhell Luzhnica, Ben Day, and Pietro Lio.
\newblock Clique pooling for graph classification.
\newblock In \emph{ICLR Workshop}, 2019.

\bibitem[McInnes et~al.(2018)McInnes, Healy, and Melville]{mcinnes2018umap}
Leland McInnes, John Healy, and James Melville.
\newblock Umap: Uniform manifold approximation and projection for dimension reduction.
\newblock \emph{arXiv preprint arXiv:1802.03426}, 2018.

\bibitem[Ng et~al.(2001)Ng, Jordan, and Weiss]{ng2001spectral}
Andrew Ng, Michael Jordan, and Yair Weiss.
\newblock On spectral clustering: Analysis and an algorithm.
\newblock In \emph{NeurIPS}, 2001.

\bibitem[Noguchi and Schechter(1981)]{noguchi1981intracellular}
Constance~Tom Noguchi and Alan~N Schechter.
\newblock The intracellular polymerization of sickle hemoglobin and its relevance to sickle cell disease.
\newblock \emph{Blood}, 58\penalty0 (6):\penalty0 1057--1068, 1981.

\bibitem[Orth(1979)]{orth1979adrenocorticotropic}
DAVID~N Orth.
\newblock Adrenocorticotropic hormone (acth).
\newblock \emph{Methods of hormone radioimmunoassay}, 2:\penalty0 245--278, 1979.

\bibitem[{\"O}zt{\"u}rk et~al.(2018){\"O}zt{\"u}rk, {\"O}zg{\"u}r, and Ozkirimli]{ozturk2018deepdta}
Hakime {\"O}zt{\"u}rk, Arzucan {\"O}zg{\"u}r, and Elif Ozkirimli.
\newblock {DeepDTA}: deep drug--target binding affinity prediction.
\newblock \emph{Bioinformatics}, 34\penalty0 (17):\penalty0 i821--i829, 2018.

\bibitem[Pauling et~al.(1951)Pauling, Corey, and Branson]{pauling1951structure}
Linus Pauling, Robert~B Corey, and Herman~R Branson.
\newblock The structure of proteins: two hydrogen-bonded helical configurations of the polypeptide chain.
\newblock \emph{Proceedings of the National Academy of Sciences}, 37\penalty0 (4):\penalty0 205--211, 1951.

\bibitem[Qi et~al.(2017)Qi, Yi, Su, and Guibas]{qi2017pointnet}
Charles~Ruizhongtai Qi, Li Yi, Hao Su, and Leonidas~J Guibas.
\newblock Pointnet++: Deep hierarchical feature learning on point sets in a metric space.
\newblock In \emph{NeurIPS}, 2017.

\bibitem[Quan et~al.(2021)Quan, Wu, Yu, and Yang]{quan2021progressive}
Ruijie Quan, Yu Wu, Xin Yu, and Yi Yang.
\newblock Progressive transfer learning for face anti-spoofing.
\newblock \emph{IEEE TIP}, 30:\penalty0 3946--3955, 2021.

\bibitem[Rao et~al.(2019)Rao, Bhattacharya, Thomas, Duan, Chen, Canny, Abbeel, and Song]{tape2019}
Roshan Rao, Nicholas Bhattacharya, Neil Thomas, Yan Duan, Xi Chen, John Canny, Pieter Abbeel, and Yun~S Song.
\newblock Evaluating protein transfer learning with tape.
\newblock In \emph{NeurIPS}, 2019.

\bibitem[Reynolds et~al.(2009)]{reynolds2009gaussian}
Douglas~A Reynolds et~al.
\newblock Gaussian mixture models.
\newblock \emph{Encyclopedia of biometrics}, 741\penalty0 (659-663), 2009.

\bibitem[Rives et~al.(2021)Rives, Meier, Sercu, Goyal, Lin, Liu, Guo, Ott, Zitnick, Ma, et~al.]{rives2021biological}
Alexander Rives, Joshua Meier, Tom Sercu, Siddharth Goyal, Zeming Lin, Jason Liu, Demi Guo, Myle Ott, C~Lawrence Zitnick, Jerry Ma, et~al.
\newblock Biological structure and function emerge from scaling unsupervised learning to 250 million protein sequences.
\newblock In \emph{National Academy of Sciences}, 2021.

\bibitem[Sanger(1952)]{sanger1952arrangement}
Frederick Sanger.
\newblock The arrangement of amino acids in proteins.
\newblock In \emph{Advances in Protein Chemistry}, pages 1--67, 1952.

\bibitem[Sanger and Tuppy(1951)]{sanger1951amino}
Frederick Sanger and Hans Tuppy.
\newblock The amino-acid sequence in the phenylalanyl chain of insulin. 1. the identification of lower peptides from partial hydrolysates.
\newblock \emph{Biochemical journal}, 49\penalty0 (4):\penalty0 463, 1951.

\bibitem[Satorras et~al.(2021)Satorras, Hoogeboom, and Welling]{satorras2021n}
V{\i}ctor~Garcia Satorras, Emiel Hoogeboom, and Max Welling.
\newblock E (n) equivariant graph neural networks.
\newblock In \emph{ICML}, 2021.

\bibitem[Scarselli et~al.(2008)Scarselli, Gori, Tsoi, Hagenbuchner, and Monfardini]{scarselli2008graph}
Franco Scarselli, Marco Gori, Ah~Chung Tsoi, Markus Hagenbuchner, and Gabriele Monfardini.
\newblock The graph neural network model.
\newblock \emph{IEEE Transactions on Neural Networks}, 20\penalty0 (1):\penalty0 61--80, 2008.

\bibitem[Strokach et~al.(2020)Strokach, Becerra, Corbi-Verge, Perez-Riba, and Kim]{strokach2020fast}
Alexey Strokach, David Becerra, Carles Corbi-Verge, Albert Perez-Riba, and Philip~M Kim.
\newblock Fast and flexible protein design using deep graph neural networks.
\newblock \emph{Cell Systems}, 11\penalty0 (4):\penalty0 402--411, 2020.

\bibitem[Subakti et~al.(2022)Subakti, Murfi, and Hariadi]{subakti2022performance}
Alvin Subakti, Hendri Murfi, and Nora Hariadi.
\newblock The performance of bert as data representation of text clustering.
\newblock \emph{Journal of Big Data}, 9\penalty0 (1):\penalty0 1--21, 2022.

\bibitem[Sun et~al.(2004)Sun, Foster, and Boyington]{sun2004overview}
Peter~D Sun, Christine~E Foster, and Jeffrey~C Boyington.
\newblock Overview of protein structural and functional folds.
\newblock \emph{Current Protocols in Protein Science}, 35\penalty0 (1):\penalty0 17--1, 2004.

\bibitem[Townshend et~al.(2019)Townshend, Bedi, Suriana, and Dror]{townshend2019end}
Raphael Townshend, Rishi Bedi, Patricia Suriana, and Ron Dror.
\newblock End-to-end learning on 3d protein structure for interface prediction.
\newblock In \emph{NeurIPS}, 2019.

\bibitem[Tsubaki et~al.(2019)Tsubaki, Tomii, and Sese]{tsubaki2019compound}
Masashi Tsubaki, Kentaro Tomii, and Jun Sese.
\newblock Compound--protein interaction prediction with end-to-end learning of neural networks for graphs and sequences.
\newblock \emph{Bioinformatics}, 35\penalty0 (2):\penalty0 309--318, 2019.

\bibitem[Vaswani et~al.(2017)Vaswani, Shazeer, Parmar, Uszkoreit, Jones, Gomez, Kaiser, and Polosukhin]{vaswani2017attention}
Ashish Vaswani, Noam Shazeer, Niki Parmar, Jakob Uszkoreit, Llion Jones, Aidan~N Gomez, {\L}ukasz Kaiser, and Illia Polosukhin.
\newblock Attention is all you need.
\newblock In \emph{NeurIPS}, 2017.

\bibitem[Veli{\v{c}}kovi{\'{c}} et~al.(2018)Veli{\v{c}}kovi{\'{c}}, Cucurull, Casanova, Romero, Li{\`{o}}, and Bengio]{velickovic2018graph}
Petar Veli{\v{c}}kovi{\'{c}}, Guillem Cucurull, Arantxa Casanova, Adriana Romero, Pietro Li{\`{o}}, and Yoshua Bengio.
\newblock Graph attention networks.
\newblock In \emph{ICLR}, 2018.

\bibitem[Wang et~al.(2024)Wang, Fan, Quan, and Yang]{wang2024protchatgpt}
Chao Wang, Hehe Fan, Ruijie Quan, and Yi Yang.
\newblock Protchatgpt: Towards understanding proteins with large language models.
\newblock \emph{arXiv preprint arXiv:2402.09649}, 2024.

\bibitem[Wang et~al.(2023{\natexlab{a}})Wang, Liu, Liu, Kurtin, and Ji]{wanglearning2023}
Limei Wang, Haoran Liu, Yi Liu, Jerry Kurtin, and Shuiwang Ji.
\newblock Learning hierarchical protein representations via complete 3d graph networks.
\newblock In \emph{ICLR}, 2023{\natexlab{a}}.

\bibitem[Wang et~al.(2018)Wang, Shen, Porikli, and Yang]{wang2018semi}
Wenguan Wang, Jianbing Shen, Fatih Porikli, and Ruigang Yang.
\newblock Semi-supervised video object segmentation with super-trajectories.
\newblock \emph{IEEE TPAMI}, 41\penalty0 (4):\penalty0 985--998, 2018.

\bibitem[Wang et~al.(2023{\natexlab{b}})Wang, Han, Zhou, and Liu]{wang2022visual}
Wenguan Wang, Cheng Han, Tianfei Zhou, and Dongfang Liu.
\newblock Visual recognition with deep nearest centroids.
\newblock In \emph{ICLR}, 2023{\natexlab{b}}.

\bibitem[Wang and Ji(2020)]{wang2020second}
Zhengyang Wang and Shuiwang Ji.
\newblock Second-order pooling for graph neural networks.
\newblock \emph{IEEE TPAMI}, 2020.

\bibitem[Wang et~al.(2022)Wang, Combs, Brand, Calvo, Xu, Price, Golovach, Salawu, Wise, Ponnapalli, et~al.]{wang2022lm}
Zichen Wang, Steven~A Combs, Ryan Brand, Miguel~Romero Calvo, Panpan Xu, George Price, Nataliya Golovach, Emmanuel~O Salawu, Colby~J Wise, Sri~Priya Ponnapalli, et~al.
\newblock Lm-gvp: an extensible sequence and structure informed deep learning framework for protein property prediction.
\newblock \emph{Scientific Reports}, 12\penalty0 (1):\penalty0 6832, 2022.

\bibitem[Wheatland(2004)]{wheatland2004molecular}
R Wheatland.
\newblock Molecular mimicry of acth in sars--implications for corticosteroid treatment and prophylaxis.
\newblock \emph{Medical Hypotheses}, 63\penalty0 (5):\penalty0 855--862, 2004.

\bibitem[Xu et~al.(2022)Xu, De~Mello, Liu, Byeon, Breuel, Kautz, and Wang]{xu2022groupvit}
Jiarui Xu, Shalini De~Mello, Sifei Liu, Wonmin Byeon, Thomas Breuel, Jan Kautz, and Xiaolong Wang.
\newblock Groupvit: Semantic segmentation emerges from text supervision.
\newblock In \emph{CVPR}, 2022.

\bibitem[Xu et~al.(2019)Xu, Hu, Leskovec, and Jegelka]{xu2018powerful}
Keyulu Xu, Weihua Hu, Jure Leskovec, and Stefanie Jegelka.
\newblock How powerful are graph neural networks?
\newblock In \emph{ICLR}, 2019.

\bibitem[Yang et~al.(2018)Yang, Wu, Bedbrook, and Arnold]{yang2018learned}
Kevin~K Yang, Zachary Wu, Claire~N Bedbrook, and Frances~H Arnold.
\newblock Learned protein embeddings for machine learning.
\newblock \emph{Bioinformatics}, 34\penalty0 (15):\penalty0 2642--2648, 2018.

\bibitem[Yin et~al.(2022)Yin, Zhou, Zhang, Fang, Xu, Shen, and Wang]{yin2022proposalcontrast}
Junbo Yin, Dingfu Zhou, Liangjun Zhang, Jin Fang, Cheng-Zhong Xu, Jianbing Shen, and Wenguan Wang.
\newblock Proposalcontrast: Unsupervised pre-training for lidar-based 3d object detection.
\newblock In \emph{ECCV}, 2022.

\bibitem[Ying et~al.(2018)Ying, You, Morris, Ren, Hamilton, and Leskovec]{ying2018hierarchical}
Zhitao Ying, Jiaxuan You, Christopher Morris, Xiang Ren, Will Hamilton, and Jure Leskovec.
\newblock Hierarchical graph representation learning with differentiable pooling.
\newblock In \emph{NeurIPS}, 2018.

\bibitem[Zhan et~al.(2020)Zhan, Xie, Liu, Ong, and Loy]{zhan2020online}
Xiaohang Zhan, Jiahao Xie, Ziwei Liu, Yew-Soon Ong, and Chen~Change Loy.
\newblock Online deep clustering for unsupervised representation learning.
\newblock In \emph{CVPR}, 2020.

\bibitem[Zhang et~al.(2018)Zhang, Cui, Neumann, and Chen]{zhang2018end}
Muhan Zhang, Zhicheng Cui, Marion Neumann, and Yixin Chen.
\newblock An end-to-end deep learning architecture for graph classification.
\newblock In \emph{AAAI}, 2018.

\bibitem[Zhang et~al.(2023)Zhang, Xu, Jamasb, Chenthamarakshan, Lozano, Das, and Tang]{zhang2022protein}
Zuobai Zhang, Minghao Xu, Arian Jamasb, Vijil Chenthamarakshan, Aurelie Lozano, Payel Das, and Jian Tang.
\newblock Protein representation learning by geometric structure pretraining.
\newblock In \emph{ICLR}, 2023.

\bibitem[Zhou et~al.(2022)Zhou, Wang, Konukoglu, and Van~Gool]{zhou2022rethinking}
Tianfei Zhou, Wenguan Wang, Ender Konukoglu, and Luc Van~Gool.
\newblock Rethinking semantic segmentation: A prototype view.
\newblock In \emph{CVPR}, 2022.

\end{thebibliography}
}

\clearpage
\setcounter{page}{1}

\appendix
\centerline{\maketitle{\textbf{SUMMARY OF THE APPENDIX}}}

This appendix contains additional details for the CVPR 2024 paper, titled \textit{``Clustering for Protein Representation Learning"}. 
This appendix provides more details of our approach, additional literature review, further discussions, additional experimental results, broader impacts, and limitations. These topics are organized as follows:
\begin{itemize}
	\item \S\ref{sec:details_training}: Details of Training Setup
	\item \S\ref{sec:evaluation_metrics}: Details of Evaluation Metrics
	\item \S\ref{subsec:rotate}: Rotation Invariance
	\item \S\ref{sec:algorithm}: Clustering Algorithm	
	\item \S\ref{sec:more_review}: Additional Literature Review
	\item \S\ref{sec:more_quantitative}: More Quantitative Results
	\item \S\ref{sec:more_vis}: More Qualitative Results
	\item \S\ref{sec:social_impact}: Broader Impacts
	\item \S\ref{sec:limitations}: Limitations

\end{itemize}

\section{Details of Training Setup}\label{sec:details_training}

In our experiments, we train our framework using a SGD optimizer with learning rate of 1e-3 and a weight decay of 5e-4.
Due to memory constraints, we set the batch size to 32, 24, 8, and 8 for EC, GO, Fold Classification, and Reaction tasks, respectively.
The framework comprises four iterations of clustering, which employ varying numbers of channels at each iteration. 
For Fold Classification, we use 256, 512, 1024 and 2048 channels for the four iterations, respectively. 
For EC, GO and Reaction, we use 128, 256, 512 and 1024 channels for the four iterations, respectively. The training process spans 200 or 300 epochs for each dataset, and the best model is selected based on the validation performance.
 Details can be seen in Table~\ref{tab:hyperpar}.

In addition, we adopt data augmentation techniques, similar to those used in~\cite{hermosilla2020intrinsic,fancontinuous} to augment the data for fold and reaction classification tasks. Specifically, we apply Gaussian noise with a standard deviation of 0.1 and anisotropic scaling within the range of [0.9, 1.1] to the amino acid coordinates in the input data. We also add the same noise to the atomic coordinates within the same amino acid to ensure that the internal structure of each amino acid remains unchanged.

\begin{table}[!h]
    \centering
    \caption{The hyperparameter configurations of our method vary across different tasks. We choose all the hyperparameters based on their performance on the validation set. See details in \S\ref{sec:details_training}.}
    \label{tab:hyperparameter}
\resizebox{1\linewidth}{!}{
			\setlength\tabcolsep{0pt}
			\renewcommand\arraystretch{1.0}
        \begin{tabular}{lcccccc}
            \thickhline
            \multicolumn{1}{l}{{Hyperparameter}}
            & {EC} & {GO-BP} & {GO-MF} & {GO-CC} & {Fold Classification} & {Reaction} \\
            \thickhline
            batch size & 32 & 24 & 24 & 24   & 8 & 8\\
            Channels    &  [128,256,512,1024] & [128,256,512,1024]  &   [128,256,512,1024] & [128,256,512,1024]  & [256,512,1024,2048] & [128,256,512,1024] \\
            \# epoch & 300 & 300 & 300 & 300   & 200 & 200\\
            
            \thickhline
        \end{tabular}}
        \label{tab:hyperpar}
\end{table}
	
\section{Details of Evaluation Metrics}\label{sec:evaluation_metrics}
We first present the details of evaluation metrics for enzyme commission number prediction and gene ontology term prediction.
The objective of these tasks is to determine whether a protein possesses specific functions, which can be viewed as multiple binary classification tasks.
We define the first metric as the protein-centric maximum \textit{F}-score (\textit{F}\textsubscript{max}). This score is obtained by calculating the precision and recall for each protein and then averaging the scores over all proteins.
To be more specific, for a given target protein $i$ and a decision threshold $\lambda\in[0,1]$, we compute the precision and recall as follows:
\begin{equation}\nonumber\small
\begin{aligned}
    \text{precision}_i(\lambda)=\frac{\sum_{a} \mathds{1}[a\in P_i(\lambda) \cap G_i]}{\sum_{a} \mathds{1}[a\in P_i(\lambda)]}, \\
    \text{recall}_i(\lambda)=\frac{\sum_{a} \mathds{1}[a\in P_i(\lambda) \cap G_i]}{\sum_{a} \mathds{1}[a\in G_i]},
  \end{aligned}
\end{equation}
where $a$ represents a function term in the ontology, $G_i$ is a set of experimentally determined function terms for protein $i$, $P_i(\lambda)$ denotes the set of predicted terms for protein $i$ with scores greater than or equal to $\lambda$, and $\mathds{1}[\cdot]\in\{0,1\}$ is an indicator function that is equal to $1$ if the condition is true.

Then, the average precision and recall over all proteins at threshold $\lambda$ is defined as:
\begin{equation}\nonumber\small
\begin{aligned}
    \text{precision}(\lambda)=\frac{\sum\nolimits_{i} \text{precision}_i(\lambda)}{M(\lambda)}, \\
    \text{recall}(\lambda)=\frac{\sum\nolimits_{i} \text{recall}_i(\lambda)}{N},
\end{aligned}
\end{equation}
where we use $N$ to represent the number of proteins, and $M(\lambda)$ to denote the number of proteins on which at least one prediction was made above threshold $\lambda$, \emph{i.e.}, $|P_i(\lambda)|>0$.
By combining these two measures, the maximum F-score is defined as the maximum \textit{F}-measure value obtained across all thresholds:
\begin{equation}\nonumber\small
    \text{\textit{F}\textsubscript{max}}=
    \max_{\lambda}\left\{\frac{2\times \text{precision}(\lambda)\times \text{recall}(\lambda)}{\text{precision}(\lambda)+ \text{recall}(\lambda)}\right\}.
\end{equation}

The\ms second\ms metric,\ms mean\ms accuracy,\ms is\ms calculated\ms as\ms the\ms average\ms precision\ms scores\ms for\ms all\ms protein-function\ms pairs,\ms which is equivalent to the micro average precision score for multiple binary classification.

\section{Rotation Invariance}
\textbf{Rotation Invariance}.\label{subsec:rotate}
To make our method rotationally invariant~\cite{satorras2021n},
we augment  the distance information ${d}_k^n$ by using a relative spatial encoding~\cite{ingraham2019generative}:
\begin{equation}\small
\label{eq:relaspa}
	{d}_k^n=\big(d(||\bm{z}_k-\bm{z}_n||),
	~~\bm{O}_n^\top{\frac{\bm{z}_k-\bm{z}_n}
	{||\bm{z}_k-\bm{z}_n||}},
	~~q(\bm{O}_n^\top\bm{O}_k^n)\big),
\end{equation}
where
${\bm{O}_n}=[\bm{b}_n,~\bm{j}_n,~\bm{b}_n\times{\bm{j}_n}]$, $\bm{b}_n$
denotes the negative bisector of angle between the ray $(v_{n-1}-v_n)$ and $(v_{n+1}-v_n)$, and $\bm{j}_n$ is a unit vector normal to that plane. 
Formally, we have $\bm{u}_n=\frac{\bm{z}_n-\bm{z}_{n-1}}{||\bm{z}_n-\bm{z}_{n-1}||}\in{\mathbb{R}^{3}}$, $\bm{b}_n=\frac{\bm{u}_n-\bm{u}_{n+1}}{||\bm{u}_n-\bm{u}_{n+1}||}\in{\mathbb{R}^{3}}$, $\bm{j}_n=\frac{\bm{u}_n\times\bm{u}_{n+1}}{||\bm{u}_n\times\bm{u}_{n+1}||}\in{\mathbb{R}^{3}}$, where $\times$ is the cross product.
The first term in Eq.~\ref{eq:relaspa} is a distance encoding ${d}(\cdot)$ lifted into the radius $r$, the second term is a direction encoding that corresponds to the relative direction of $v_k^n\rightarrow{v_n}$, and the third term is an orientation encoding ${q}(\cdot)$ of the quaternion representation of the spatial rotation matrix.
This encoding approach allows us to capture both local and global geometric information while being invariant to different orientations. Related experimental results are seen in Table~\ref{tab:aba_roi_app}.

\vspace{-10pt}
\section{Neural Clustering Algorithm}\label{sec:algorithm}
\vspace{-10pt}
\begin{algorithm}[h]
		\caption{Pseudo-code of neural clustering}
		\label{alg:cluster_algorithm}
		\SetKwInOut{Input}{Input}\SetKwInOut{Output}{Output}\SetKwInOut{Intermediate}{Intermediate}
		\Input{~Protein $\mathcal{P}=(\mathcal{V},\mathcal{E},\mathcal{Y})$; Amino acid embedding $\bm{e}_j$ for amino acid $v_j\in{\mathcal{V}}$; Cluster nomination ratio $\omega$; Nomination operator \textsc{Nominate}; Index selection operator \textsc{IndexSelect};  Add self-loop operator \textsc{addselfloops}; Spherosome clustering operator \textsc{RADIUS}; Spherosome clustering radius $r$; ReLU activation function $\sigma$; Geometric coordinates $Pos$; Geometric orientations $Ori$; Sequential orders $Seq$}
		\Intermediate{~Clustered features $X^c$ and Scored cluster features $\hat{X}^c$; Adjacency matrix $A$; Edge index $E$; Cluster scores vector $\Phi$; Nominated index $index$}
		\Output{~Nominated amino acid features $X$, coordinates $Pos$, orientations $Ori$, sequential orders $Seq$}
		\BlankLine
		\For{$t=1,2,3,4$}
		{
		        $A \leftarrow \textsc{Radius}(Pos, r)$\;
		        $E \leftarrow \textsc{AddSelfLoops}(A)$\;

		\For{$n=1...N_{t-1}$}
		{
		    $\tilde{\bm{x}}_{n} \leftarrow \Vec{0}$\;
			\For{$k=1...K$}
			{
				$\bm{g}_k^n, \bm{o}_k^n, {d}_k^n, {s}_k \leftarrow (Pos, Ori, Seq)$\;
				$\bm{x}_k^n \leftarrow f(\bm{g}_k^n, \bm{o}_k^n, {d}_k^n, {s}_k, \bm{e}_k)$\;
				$\gamma^{n}_{k} \leftarrow softmax(\sigma([\bm{W}_{1}\bm{x}_n,\bm{x}_k^n]))$\;
				$\tilde{\bm{x}}_{n} \leftarrow \tilde{\bm{x}}_{n} + \gamma^{n}_{k} \bm{x}_k^n$\;
			}
			$X^c_n \leftarrow \tilde{\bm{x}}_{n}$\;
		}

		$\Phi \leftarrow \textsc{GCN}(X^c, E)$\;
		$\hat{X}^c \leftarrow \Phi \odot X^c $\;
		${N_t} \leftarrow \lfloor\omega\cdot{N_{t-1}}\rfloor$\;
		$index \leftarrow \textsc{Nominate}(\Phi,  N_t)$\;
		$X, Pos, Ori, Seq \leftarrow \textsc{IndexSelect}(\hat{X}^c, Pos, Ori, Seq, index)$\;
		}

	\end{algorithm}

\section{Additional Literature Review}\label{sec:more_review}
\textbf{Graph Pooling.} 
Graph pooling designs have been proposed to achieve a useful and rational graph representation. These designs can be broadly categorized into two types~\cite{liu2022graph}: Flat Pooling~\cite{duvenaud2015convolutional,xu2018powerful,zhang2018end,wang2020second,baek2021accurate} and Hierarchical Pooling~\cite{ying2018hierarchical,luzhnica2019clique,ali2023quasi,lee2019self,gao2019graph,gao2021ipool,gao2021topology}. Flat Pooling generates a graph-level representation in a single step by primarily calculating the average or sum of all node embeddings without consideration of the intrinsic hierarchical structures
of graphs, which causes information loss~\cite{knyazev2019understanding}. On the other hand, Hierarchical Pooling gradually reduces the size of the graph.
Previous graph pooling algorithms, as variants of GCN, still follow the message passing pipeline. Typically, they are hard to be trained and need many extra regularizations and/or operations. For example, 
Diffpool~\cite{ying2018hierarchical} is trained with an auxiliary link prediction objective. Besides, it generates a dense assignment matrix thus incurring a quadratic storage complexity. 
Top-K pooling~\cite{gao2019graph}, adopts a Unet-like, graph encoder-decoder architecture, which is much more complicated than our model but only learns a simple scalar projection score for each node.

In contrast, our clustering-based algorithm is more principled and elegant. It can address the sparsity concerns of Diffpool and capture rich protein structure information by aggregating amino acids to form clusters instead of learning from a single node. Furthermore, it sticks to the principle of clustering throughout its algorithmic design: SCI step is to form the clusters by considering geometrical relations among amino acids; CRE step aims to extract cluster-level representations; CN step is for the selection of important cluster centers. It essentially combines unsupervised clustering with supervised classification. The forward process of our model is inherently a neural clustering process, which is more transparent and without any extra supervision.

\section{More Quantitative Results}\label{sec:more_quantitative}

\begin{table*}[h]
  \centering
  \caption{Analysis of the impact of rotation invariance and different numbers of CRE blocks (\S\ref{subsec:analysis}).}
    \label{tab:aba_roi_app}
    \resizebox{\textwidth}{!}{
		\setlength\tabcolsep{14pt}
		\renewcommand\arraystretch{1}
        \begin{tabular}{c||c||c|ccc|cccc|c}
            \thickhline 
            \rowcolor{mygray}{Rotation} & &
            \multirow{2}{*}{ } &
           
            \multicolumn{3}{c|}{ {GO}}& 
            \multicolumn{4}{c|}{ {Fold Classification}}&
            \multirow{2}{*}{ }\\
            \cline{4-6}
            \cline{7-10}
            \rowcolor{mygray}{Invariant}& \multirow{-2}{*}{$B$}  & \multirow{-2}{*}{EC} & {BP} &  {MF}&  {CC} &   {Fold} &  {Super.} &  {Fam.} &  {Avg.} & \multirow{-2}{*}{Reaction}\\
            \hline \hline
   \ding{52} & \textit{1} 
             & 0.825 & 0.430 &  0.618 & 0.464 &  57.7 & 76.3  & 99.4  & 77.8  & 87.6 \\
    \ding{52} & \textit{2} 
             & \textbf{0.866} & \textbf{0.474} & \textbf{0.675} & \textbf{0.483} & \textbf{63.1} &  \textbf{81.2}  &  \textbf{99.6} &  \textbf{81.3} & \textbf{89.6}   \\
   \ding{52} & \textit{3}
             & 0.857 & 0.466 &  0.669 & 0.474 &  61.8 & 80.2  & 99.5  & 80.5  & 88.9 \\
   \ding{56}   & \textit{2} 
             & 0.781 & 0.392 & 0.614 & 0.436 & 56.4  &  75.3  & 97.9  & 76.4  & 87.1 \\

            \hline
        \end{tabular}} 
\end{table*}

\noindent
\textbf{Rotation Invariance.} 
We compare our framework with and without rotation invariance on all four tasks. The results are shown in Table~\ref{tab:aba_roi_app}, where we can see that rotation invariance improves the performance of our framework on all tasks. For example, on protein fold classification, rotation invariance\ms boosts\ms the\ms average\ms accuracy\ms from\ms 76.4\% to\ms 81.3\%.\ms This\ms indicates\ms that\ms rotation\ms invariance\ms can\ms help\ms our\ms framework\ms to\ms capture\ms the\ms geometric\ms information\ms of\ms proteins\ms more\ms effectively\ms and\ms robustly.

\noindent
\textbf{Number of CRE Blocks.}
In our framework, we use $B$ CRE blocks at each clustering iteration to extract cluster features. We study the impact of using different values of $B$ from $1$ to $3$ on all four sub-tasks. We stop using $B\!>\!3$ as the required memory exceeds the computational limit of our hardware. The results are shown in Table~\ref{tab:aba_roi_app}, where we can find that $B\!=\!2$ achieves the best performance on all tasks. For instance, on enzyme reaction classification, $\!B=\!2$ achieves an accuracy of 89.6\%, while if $B\!=\!1$ or $B\!=\!3$,  the accuracy drops to 87.6\% and 88.9\%, respectively. This suggests that using two CRE blocks is sufficient to capture the cluster information and adding more blocks does not bring significant improvement but increases the computational cost.

\section{More Qualitative Results}\label{sec:more_vis}

\begin{figure*}[t]
    \centering
    \includegraphics[width=\textwidth]{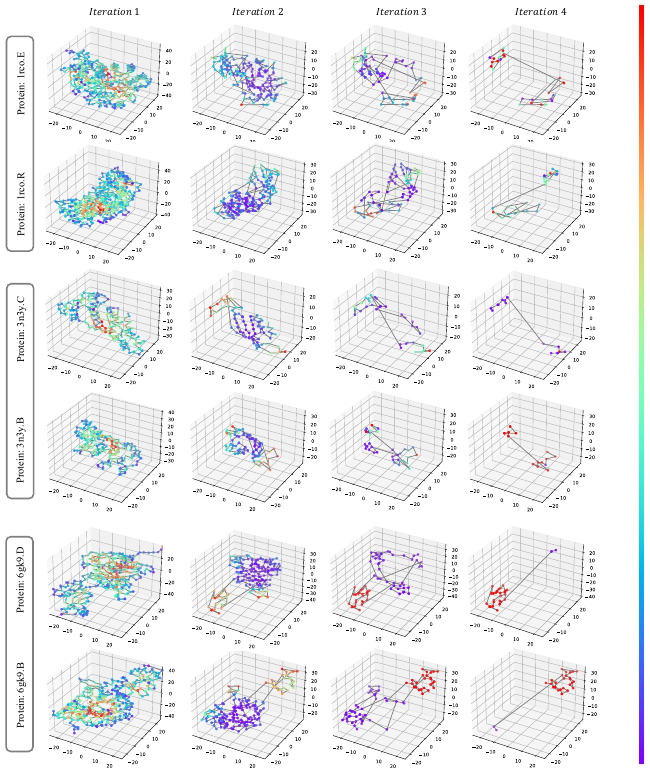}
    \caption{More visualization results. See related analysis in \S\ref{sec:more_vis}.}
    \vspace{-10pt}
    \label{fig:more_vis}
\end{figure*}

More visualization results of the clustering results are presented in Fig.~\ref{fig:more_vis}. The color of the node denotes the score calculated in our CN step. 
For example, in the first row, we can see that the protein `1rco.E' has a helical structure with some loops. After the first iteration of clustering, our method selects some amino acids that are located at the ends or bends of these loops and helices as the center nodes for the next iteration. 
These amino acids may play an important role in stabilizing the protein structure or interacting with other molecules. After the second iteration of clustering, our method further narrows down the number of center nodes by selecting those that have high scores. 
These amino acids may form functional domains or motifs that are essential for the protein function. Our method finally identifies a few amino acids that have the highest scores and are most representative of the protein structure and function. 
Through visualizing the clustering results at each iteration, we can explicitly understand how our method progressively discovers the critical components of different proteins 
by capturing their structural  features in a hierarchical way.

 In addition, as shown in the figure, we present pairs of protein chains from the same family or same protein (\ie, `1rco.R' and `1rco.E', `3n3y.B' and `3n3y.C', `6gk9.B' and `6gk9.D'). For example, 
 we observe that the clustering results of `3n3y.B' and `3n3y.C' (two chains of the same protein) are very similar, indicating that they have similar critical amino acids that determine their structure and function. This observation is consistent with the biological reality that proteins from the same family or same protein often have long stretches of similar amino acid sequences within their primary structure, suggesting  that our method is effective in identifying these critical amino acids.

\section{Broader Impacts}\label{sec:social_impact}
Our neural clustering framework for protein representation learning has several potential applications and implications for society. Protein representation learning can help advance our understanding of protein structure and function, which are essential for many biological processes and diseases. By discovering the critical components of proteins, our method can inspire protein design, engineering, and modification, which can lead to the development of novel therapies, drugs, and biotechnologies. For example, our framework can assist in designing new protein sequences that exhibit specific properties or functions, such as catalyzing biochemical reactions or binding to other molecules. This can enable the creation of new enzymes, antibodies, vaccines, and biosensors that can have a positive impact on human health and well-being.

\section{Limitations}\label{sec:limitations}
Our neural clustering framework for protein representation learning has several limitations that need to be addressed in future work. First, our framework relies on the availability of protein structures, which are not always easy to obtain or predict. Although our framework can leverage both sequence-based and structure-based features, it may lose some information that is only encoded in the 3D structure. Second, our framework assumes that the critical components of a protein are determined by its amino acid sequence and structure, but it does not consider other factors that may affect protein function, such as post-translational modifications, interactions with other molecules, or environmental conditions. Third, our framework does not explicitly account for the evolutionary relationships among proteins, which may provide useful information for protein representation learning. Incorporating phylogenetic information into our framework may enhance its ability to capture the functional diversity and similarity of proteins.

\end{document}